# A General Approach to Belief Change
# in Answer Set Programming

JAMES DELGRANDE

Simon Fraser University

TORSTEN SCHAUB

Universität Potsdam

and

HANS TOMPITS and STEFAN WOLTRAN

Technische Universität Wien

---

We address the problem of belief change in (nonmonotonic) logic programming under answer set semantics. Unlike previous approaches to belief change in logic programming, our formal techniques are analogous to those of distance-based belief revision in propositional logic. In developing our results, we build upon the model theory of logic programs furnished by SE models. Since SE models provide a formal, monotonic characterisation of logic programs, we can adapt techniques from the area of belief revision to belief change in logic programs.

We first consider belief revision: given logic programs $P$ and $Q$, the goal is to determine a program $R$ that corresponds to the revision of $P$ by $Q$, denoted $P * Q$. We investigate several specific operators, including (logic program) expansion and two revision operators based on the distance between the SE models of logic programs. It proves to be the case that expansion is an interesting operator in its own right, unlike in classical AGM-style belief revision where it is relatively uninteresting. Expansion and revision are shown to satisfy a suite of interesting properties; in particular, our revision operators satisfy the majority of the AGM postulates for revision.

Second, we consider approaches for merging logic programs. Given logic programs $P_1, \ldots, P_n$, the goal is to provide characterisations of the merging of these programs. Again, our formal techniques are based on notions of relative distance between the underlying SE models of the logic programs. Two approaches are examined. The first informally selects those models of the programs that vary the least from the models of the other programs. The second approach informally selects those models of a program $P_0$ that are closest to the models of programs $P_1, \ldots, P_n$. In this case, $P_0$ can be thought of as analogous to a set of database integrity constraints. We examine properties of these operators with regards to how they satisfy relevant postulate sets.

We also present encodings for computing the revision as well as the merging of logic programs within the same logic programming framework, giving rise to a direct implementation of our approach in terms of off-the-shelf answer set solvers. These encodings reflect in turn the fact that our change operators do not increase the complexity of the base formalism.

---

The first author was supported by a Canadian NSERC Discovery Grant. The second author was supported by the German Science Foundation (DFG) under grant SCHA 550/8-1. The third author was supported by the Austrian Science Fund (FWF) under project P21698. Authors' address: J. Delgrande, Simon Fraser University, Burnaby, B.C., Canada, V5A 1S6, e-mail: jim@cs.sfu.ca. T. Schaub, Universität Potsdam, August-Bebel-Straße 89, D-14482 Potsdam, Germany, e-mail: torsten@cs.uni-potsdam.de. H. Tompits and S. Woltran, Technische Universität Wien, Favoritenstraße 9-11, A-1040 Vienna, Austria, e-mail: tompits@kr.tuwien.ac.at, woltran@dbai.tuwien.ac.at.

## 1. INTRODUCTION

*Answer set programming* (ASP) [Gelfond and Lifschitz 1988; Baral 2003] has emerged as a major area of research in knowledge representation and reasoning (KRR). On the one hand, ASP has an elegant and conceptually simple theoretical foundation, while on the other hand efficient implementations of ASP solvers exist which have been finding applications to practical problems. However, as is the case with any large program or body of knowledge, a logic program is not a static object in general, but rather it will evolve and be subject to change, whether as a result of correcting information in the program, adding to the information already present, or in some other fashion modifying the knowledge represented in the program.

Since knowledge is continually evolving and subject to change, there is a need to be able to revise logic programs as new information is received. In KRR, the area of *belief revision* [Alchourrón et al. 1985; Gärdenfors 1988] addresses just such change to a knowledge base. In *AGM belief revision* (named after the aforecited developers of the approach) one has a knowledge base $K$ and a formula $\alpha$, and the issue is how to consistently incorporate $\alpha$ in $K$ to obtain a new knowledge base $K'$. The interesting case is when $K \cup \{\alpha\}$ is inconsistent, since beliefs have to be dropped from $K$ before $\alpha$ can be consistently added. Hence, a fundamental issue concerns how such change should be managed.

In classical propositional logic, specific belief revision operators have been proposed based on the distance between *models* of a knowledge base and a formula for revision. That is, a characterisation of the revision of a knowledge base $K$ by formula $\alpha$ is to set the models of the revised knowledge base $K'$ to be the models of $\alpha$ that are "closest" to those of $K$. Of course the notion of "closest" needs to be pinned down, but natural definitions based on the Hamming distance [Dalal 1988] and set containment with regards to propositional letters [Satoh 1988] are well known.

In addition to belief revision (along with the dual notion of belief *contraction*), a second major class of belief change operators addresses the *merging* of knowledge bases. The problem of merging multiple, potentially conflicting bodies of information arises in various different contexts. For example, an agent may receive reports from differing sources of knowledge, or from sets of sensors that need to be reconciled. As well, an increasingly common phenomenon is that collections of data may need to be combined into a coherent whole. In these cases, the problem is that of combining knowledge sets that may be jointly inconsistent in order to get a consistent set of merged beliefs. Again, as in belief revision, specific operators for merging knowledge bases have been developed based on the distance between models of the underlying knowledge bases [Baral et al. 1992; Revesz 1993; Liberatore and Schaerf 1998; Meyer 2001; Konieczny and Pino Pérez 2002; Konieczny et al. 2002].

It is natural then to consider belief change in the context of logic programs. Indeed, there has been substantial effort in developing approaches to so-called *logic program updating* under answer set semantics (we discuss previous work in the next section). Unfortunately, given the nonmonotonic nature of answer set programs, the problem of change in logic programs has appeared to be intrinsically more difficult than in a monotonic setting. In this paper, our goal is to reformulate belief change in logic programs in a manner analogous to belief change in classical propositional logic, and to investigate specific belief revision and merging operators for extended logic programs. Central for our approach are *SE models* [Turner 2003], which are semantic structures characterising *strong equivalence* between programs [Lifschitz et al. 2001]. This particular kind of equivalence plays a major role for different problems in logic programming—in particular, in program simplifications and modularisation. This is due to the fact that strong equivalence gives rise to a *substitution principle* in the sense that, for all programs $P, Q$, the programs $P \cup R$ and $Q \cup R$ have the same answer sets, for *any* program $R$. As is well known, ordinary equivalence between programs (which holds if two programs have the same answer sets) does not yield a substitution principle. Hence, strong equivalence can be seen as the logic programming analogue of ordinary equivalence in classical logic. The important aspect of strong equivalence is that it coincides with equivalence in a specific *monotonic logic*, the logic of *here and there* (HT), which is intermediate between intuitionistic logic and classical logic. As shown by Turner [2003], equivalence between programs in HT corresponds in turn to equality between sets of SE models. Details on these concepts are given in the next section; the key point is that logic programs can be expressed in terms of a non-classical but monotonic logic, and it is this point that we exploit here.

More specifically, given this monotonic characterisation (via sets of SE models) of strong equivalence, we adapt techniques for belief change in propositional logic to belief change in logic programs. Hence we come up with specific operators for belief change in ASP analogous to operators in propositional logic. We first consider an *expansion* operator. In classical logic, the expansion of knowledge base $K$ by formula $\alpha$ amounts to the deductive closure of $K \cup \{\alpha\}$. Hence it is not a very interesting operator, serving mainly as a tool for expressing concepts in belief revision and its dual, contraction. In logic programs however, expansion appears to be a more useful operator, perhaps due to the apparent "looser" notion of satisfiability provided by SE models. As well, it has appealing properties. We next develop revision operators based on notions of distance between SE models, and, following this, merging operators.

In characterising the merging of logic programs, the central idea is that the SE models of the merged program are again those that are in some sense "closest" to the SE models of the programs to be merged. However, as with merging knowledge bases expressed in classical logic, there is no single preferred notion of distance nor closeness, and consequently different approaches have been defined for combining sources of information. We introduce two merging operators for logic programs under answer set semantics. Both operators take an arbitrary (multi)set of logic programs as argument. The first operator can be regarded an instance of what Liberatore and Schaerf [1998] call *arbitration*. Basically (SE) models are selected from among the SE models of the programs to be merged; in a sense this operator is a natural extension of our belief revision operator. The second merging operator can be regarded as an instance of the one discussed by Konieczny and Pino Pérez [2002]. Here, models of a designated program (representing information analogous to database integrity constraints) are selected that are closest to (or perhaps, informally, represent the best compromise among) the models of the programs to be merged.

Notably, in our approaches there is effectively no mention of answer sets; rather definitions of expansion and revision are given entirely with respect to logic programs. Notably too, our operators are syntax independent, which is to say, they are independent of how a logic program is expressed; hence, our operators deal with the *logical content* of a logic program.

Following an introductory background section, we show that there is a ready mapping between concepts in belief revision in classical logic and in ASP; this serves to place belief revision in ASP firmly in the "standard" belief revision camp. After this we describe in Section 3 our approaches to belief expansion and revision in ASP. We then employ these techniques in the following section to address the merging of logic programs. In either case, we discuss central properties and give complexity results. Then, in Section 5, we show how we can in fact express the process of belief change in ASP itself, giving a direct way to compute our introduced belief change operators. We conclude with a discussion. Proofs of results are contained in an appendix.

## 2.  BACKGROUND AND FORMAL PRELIMINARIES

### 2.1  Answer Set Programming

2.1.1  *Syntax and Semantics.* A (*generalised*) *logic program*[1] (GLP) over an alphabet $\mathcal{A}$ is a finite set of rules of the form

$$a_1; \ldots; a_m; \sim b_{m+1}; \ldots; \sim b_n \leftarrow c_{n+1}, \ldots, c_o, \sim d_{o+1}, \ldots, \sim d_p, \qquad (1)$$

where $a_i, b_j, c_k, d_l \in \mathcal{A}$ are *atoms*, for $1 \leq i \leq m < j \leq n < k \leq o < l \leq p$. Operators ';' and ',' express disjunctive and conjunctive connectives. A *default literal* is an atom $a$ or its (default) negation $\sim a$. A rule $r$ as in (1) is called a *fact* if $p = 1$, *normal* if $n = 1$, *positive* if $m = n$ and $o = p$, *disjunctive* if $m = n$, and an *integrity constraint* if $n = 0$, yielding an empty disjunction, sometimes denoted by $\bot$ for convenience. Accordingly, a program is called *disjunctive* (or a DLP) if it consists of disjunctive rules only. Likewise, a program is *normal* (resp., *positive*) iff all rules in it are normal (resp., positive). We furthermore define the *head* and *body* of a rule, $H(r)$ and $B(r)$, by:

$$H(r) = \{a_1, \ldots, a_m, \sim b_{m+1}, \ldots, \sim b_n\} \qquad \text{and}$$
$$B(r) = \{c_{n+1}, \ldots, c_o, \sim d_{o+1}, \ldots, \sim d_p\}.$$

Moreover, given a set $X$ of literals, we define

$$X^+ = \{a \in \mathcal{A} \mid a \in X\},$$
$$X^- = \{a \in \mathcal{A} \mid \sim a \in X\}, \text{ and}$$
$$\sim X = \{\sim a \mid a \in X \cap \mathcal{A}\}.$$

For simplicity, we sometimes use a set-based notation, expressing a rule as in (1) as $H(r)^+; \sim H(r)^- \leftarrow B(r)^+, \sim B(r)^-$.

In what follows, we restrict ourselves to a finite alphabet $\mathcal{A}$. An interpretation is represented by the subset of atoms in $\mathcal{A}$ that are true in the interpretation. A (*classical*) *model* of a program $P$ is an interpretation in which all of the rules in $P$ are true according to the standard definition of truth in propositional logic, and where default negation is treated as classical negation. By $Mod(P)$ we denote the set of all classical models of $P$. An *answer set* $Y$ of a program $P$ is a subset-minimal model of

$$\{H(r)^+ \leftarrow B(r)^+ \mid r \in P, \ H(r)^- \subseteq Y, \ B(r)^- \cap Y = \emptyset\}.$$

---

[1]Such programs were first considered by Lifschitz and Woo [1992] and called *generalised disjunctive logic programs* by Inoue and Sakama [1998].

The set of all answer sets of a program $P$ is denoted by $AS(P)$. For example, the program $P = \{a \leftarrow, \quad c; d \leftarrow a, \sim b\}$ has answer sets $AS(P) = \{\{a, c\}, \{a, d\}\}$.

2.1.2 *SE Models.* As defined by Turner [2003], an *SE interpretation* is a pair $(X, Y)$ of interpretations such that $X \subseteq Y \subseteq \mathcal{A}$. An SE interpretation is an *SE model* of a program $P$ if $Y \models P$ and $X \models P^Y$. The set of all SE models of a program $P$ is denoted by $SE(P)$. Note that $Y$ is an answer set of $P$ iff $(Y, Y) \in SE(P)$ and no $(X, Y) \in SE(P)$ with $X \subset Y$ exists. Also, we have $(Y, Y) \in SE(P)$ iff $Y \in Mod(P)$.

A program $P$ is *satisfiable* just if $SE(P) \neq \emptyset$. Two programs $P$ and $Q$ are *strongly equivalent*, symbolically $P \equiv_s Q$, iff $SE(P) = SE(Q)$. Alternatively, $P \equiv_s Q$ holds iff $AS(P \cup R) = AS(Q \cup R)$, for every program $R$ [Lifschitz et al. 2001]. We also write $P \models_s Q$ iff $SE(P) \subseteq SE(Q)$. For simplicity, we often drop set-notation within SE interpretations and simply write, e.g., $(a, ab)$ instead of $(\{a\}, \{a, b\})$.

One feature of SE models is that they contain "more information" than answer sets, which makes them an appealing candidate for problems where programs are examined with respect to further extension (in fact, this is what strong equivalence is about). We illustrate this issue with the following well-known example, involving programs

$$P = \{p; q \leftarrow\} \quad \text{and} \quad Q = \left\{ \begin{array}{l} p \leftarrow \sim q \\ q \leftarrow \sim p \end{array} \right\}.$$

Here, we have $AS(P) = AS(Q) = \{\{p\}, \{q\}\}$. However, the SE models (we list them for $\mathcal{A} = \{p, q\}$) differ:

$$SE(P) = \{(p, p), (q, q), (p, pq), (q, pq), (pq, pq)\};$$
$$SE(Q) = \{(p, p), (q, q), (p, pq), (q, pq), (pq, pq), (\emptyset, pq)\}.$$

This is to be expected, since $P$ and $Q$ behave differently with respect to program extension (and thus are not strongly equivalent). Consider $R = \{p \leftarrow q, q \leftarrow p\}$. Then $AS(P \cup R) = \{\{p, q\}\}$, while $AS(Q \cup R)$ has no answer set.

A set $S$ of SE interpretations is *well-defined* if, for each $(X, Y) \in S$, also $(Y, Y) \in S$. A well-defined set $S$ of SE interpretations is *complete* if, for each $(X, Y) \in S$, also $(X, Z) \in S$, for any $Y \subseteq Z$ with $(Z, Z) \in S$. We have the following properties:

—For each GLP $P$, $SE(P)$ is well-defined.

—For each DLP $P$, $SE(P)$ is complete.

Furthermore, for each well-defined set $S$ of SE interpretations, there exists a GLP $P$ such that $SE(P) = S$, and for each complete set $S$ of SE interpretations, there exists a DLP $P$ such that $SE(P) = S$. Programs meeting these conditions can be constructed thus [Eiter et al. 2005; Cabalar and Ferraris 2007]: In case $S$ is a well-defined set of SE interpretations over a (finite) alphabet $\mathcal{A}$, define $P$ by adding

(1) the rule $r_Y : \bot \leftarrow Y, \sim(\mathcal{A} \setminus Y)$, for each $(Y, Y) \notin S$, and

(2) the rule $r_{X,Y} : (Y \setminus X); \sim Y \leftarrow X, \sim(\mathcal{A} \setminus Y)$, for each $X \subseteq Y$ such that $(X, Y) \notin S$ and $(Y, Y) \in S$.

In case $S$ is complete, define $P$ by adding

(1) the rule $r_Y$, for each $(Y, Y) \notin S$, as above, and
(2) the rule $r'_{X,Y} : (Y \setminus X) \leftarrow X, \sim(\mathcal{A} \setminus Y)$, for each $X \subseteq Y$ such that $(X, Y) \notin S$ and $(Y, Y) \in S$.

We call the resulting programs *canonical*.

For illustration, consider

$$S = \{(p, p), (q, q), (p, pq), (q, pq), (pq, pq), (\emptyset, p)\}$$

over $\mathcal{A} = \{p, q\}$. Note that $S$ is not complete. The canonical GLP is as follows:

$$
\begin{aligned}
r_\emptyset &: & \bot &\leftarrow \sim p, \sim q; \\
r_{\emptyset,q} &: & q; \sim q &\leftarrow \sim p; \\
r_{\emptyset,pq} &: p; q; \sim p; \sim q &\leftarrow .
\end{aligned}
$$

For obtaining a complete set, we have to add $(\emptyset, pq)$ to $S$. Then, the canonical DLP is as follows:

$$
\begin{aligned}
r_\emptyset &: \bot &\leftarrow \sim p, \sim q; \\
r_{\emptyset,q} &: q &\leftarrow \sim p.
\end{aligned}
$$

We conclude this subsection by introducing definitions for ordering SE models that will be needed when we come to define our belief change operators. Let $\ominus$ denote the symmetric difference operator between sets, i.e., $X \ominus Y = (X \setminus Y) \cup (Y \setminus X)$ for every set $X, Y$. We extend $\ominus$ so that it is defined for ordered pairs, as follows:

DEFINITION 1. *For every pair* $(X_1, X_2)$, $(Y_1, Y_2)$,

$$(X_1, X_2) \ominus (Y_1, Y_2) = (X_1 \ominus Y_1, X_2 \ominus Y_2).$$

Similarly, we define a notion of set containment, suitable for ordered pairs, as follows:

DEFINITION 2. *For every pair* $(X_1, X_2)$, $(Y_1, Y_2)$,

$$(X_1, X_2) \subseteq (Y_1, Y_2) \text{ iff } X_2 \subseteq Y_2, \text{ and if } X_2 = Y_2 \text{ then } X_1 \subseteq Y_1.$$

*Furthermore,* $(X_1, X_2) \subset (Y_1, Y_2)$ *iff* $(X_1, X_2) \subseteq (Y_1, Y_2)$ *and not* $(Y_1, Y_2) \subseteq (X_1, X_2)$.

As will be seen, these definitions are appropriate for SE interpretations, as they give preference to the second element of a SE interpretation.

Set cardinality is denoted as usual by $| \cdot |$. We define a cardinality-based ordering over ordered pairs of sets as follows:

DEFINITION 3. *For every pair* $(X_1, X_2)$, $(Y_1, Y_2)$,

$$|(X_1, X_2)| \leq |(Y_1, Y_2)| \text{ iff } |X_2| \leq |Y_2| \text{ and if } |X_2| = |Y_2| \text{ then } |X_1| \leq |Y_1|.$$

*Furthermore,* $|(X_1, X_2)| < |(Y_1, Y_2)|$ *iff* $|(X_1, X_2)| \leq |(Y_1, Y_2)|$ *and not* $|(Y_1, Y_2)| \leq |(X_1, X_2)|$.

As with Definition 2, this definition gives preference to the second element of an ordered pair. It can be observed that the definition yields a total preorder over ordered pairs. In the next section we return to the suitability of this definition, once our revision operators have been presented.

## 2.2 Belief Change

### 2.2.1 *Belief Revision.*

The best known and, indeed, seminal work in belief revision is the *AGM approach* [Alchourrón et al. 1985; Gärdenfors 1988], in which standards for belief *revision* and *contraction* functions are given. In the revision of a knowledge base $K$ by a formula $\phi$, the intent is that the resulting knowledge contains $\phi$, be consistent (unless $\phi$ is not), while keeping whatever information from $K$ can be "reasonably" retained. *Belief contraction* is a dual notion, in which information is removed from a knowledge base; given that it is of limited interest with respect to our approach, we do not consider it further. Moreover, it is generally accepted that a contraction function can be obtained from a revision function by the so-called *Harper identity*, and the reverse obtained via the *Levi identity*; see Gärdenfors [1988] for details. In the AGM approach it is assumed that a knowledge base is receiving information concerning a static[2] domain. Belief states are modelled by logically closed sets of sentences, called *belief sets*. A belief set is a set $K$ of sentences which satisfies the constraint

$$\text{if } K \text{ logically entails } \beta, \text{ then } \beta \in K.$$

$K$ can be seen as a partial theory of the world. For belief set $K$ and formula $\alpha$, $K + \alpha$ is the deductive closure of $K \cup \{\alpha\}$, called the *expansion* of $K$ by $\alpha$. $K_\perp$ is the inconsistent belief set (i.e., $K_\perp$ is the set of all formulas).

Subsequently, Katsuno and Mendelzon [1992] reformulated the AGM approach so that a knowledge base was represented by a formula in some language $\mathcal{L}$. The following postulates comprise Katsuno and Mendelzon's reformulation of the AGM revision postulates, where $*$ is a function from $\mathcal{L} \times \mathcal{L}$ to $\mathcal{L}$:

$(R1).$  $\psi * \mu \vdash \mu$.

$(R2).$  If $\psi \wedge \mu$ is satisfiable, then $\psi * \mu \leftrightarrow \psi \wedge \mu$.

$(R3).$  If $\mu$ is satisfiable, then $\psi * \mu$ is also satisfiable.

$(R4).$  If $\psi_1 \leftrightarrow \psi_2$ and $\mu_1 \leftrightarrow \mu_2$, then $\psi_1 * \mu_1 \leftrightarrow \psi_2 * \mu_2$.

$(R5).$  $(\psi * \mu) \wedge \phi \vdash \psi * (\mu \wedge \phi)$.

$(R6).$  If $(\psi * \mu) \wedge \phi$ is satisfiable, then $\psi * (\mu \wedge \phi) \vdash (\psi * \mu) \wedge \phi$.

Thus, revision is successful (R1), and corresponds to conjunction when the knowledge base and formula for revision are jointly consistent (R2). Revision leads to inconsistency only when the formula for revision is unsatisfiable (R3). Revision is also independent of syntactic representation (R4). Last, (R5) and (R6) express that revision by a conjunction is the same as revision by a conjunct conjoined with the other conjunct, when the result is satisfiable.

### 2.2.2 *Specific Belief Revision Operators.*

In classical belief change, the revision of a knowledge base represented by formula $\psi$ by a formula $\mu$, $\psi * \mu$, is a formula $\phi$ such that the models of $\phi$ are just those models of $\mu$ that are "closest" to those of $\psi$. There are two main specific approaches to distance-based revision. Both are related to the Hamming distance between two interpretations, that is on the set of atoms on which the interpretations disagree. The first, by Satoh [1988], is based on set containment. The second, due to Dalal [1988], uses a distance measure based on the number of atoms with differing truth values in two interpretations. A set containment-based approach seems more appropriate in the context of ASP, since answer sets are defined in terms of subset-minimal interpretations. Hence, we focus on the method of Satoh [1988], although we also consider Dalal-style revision, since it has some technical interest with respect to ASP revision.

---

[2] Note that "static" does not imply "with no mention of time". For example, one could have information in a knowledge base about the state of the world at different points in time, and revise information at these points in time.

The *Satoh revision operator*, $\psi *_s \mu$, is defined as follows. For formulas $\alpha$ and $\beta$, define $\ominus^{min}(\alpha, \beta)$ as

$$min_{\subseteq}(\{w \ominus w' \mid w \in Mod(\alpha), w' \in Mod(\beta)\}).$$

Furthermore, define $Mod(\psi *_s \mu)$ as

$$\{w \in Mod(\mu) \mid \exists w' {\in} Mod(\psi) \text{ s.t. } w \ominus w' \in \ominus^{min}(\psi, \mu)\}.$$

The *cardinality-based* or *Dalal revision operator*, $\psi *_d \mu$, is defined as follows. For formulas $\alpha$ and $\beta$, define $|\ominus|^{min}(\alpha, \beta)$ as

$$min_{\leq}(\{|w \ominus w'| \mid w \in Mod(\alpha), w' \in Mod(\beta)\}).$$

Then, $Mod(\psi *_d \mu)$ is given as

$$\{w \in Mod(\mu) \mid \exists w' {\in} Mod(\psi) \text{ s.t. } |w \ominus w'| = |\ominus|^{min}(\psi, \mu)\}.$$

2.2.3   *Belief Merging.*   Earlier work on merging operators includes approaches by Baral et al. [1992] and Revesz [1993]. The former authors propose various theory merging operators based on the selection of maximum consistent subsets in the union of the belief bases. The latter proposes an "arbitration" operator (see below) that, intuitively, selects from among the models of the belief sets being merged. Lin and Mendelzon [1999] examine *majority* merging, in which, if a plurality of knowledge bases hold $\phi$ to be true, then $\phi$ is true in the merging. Liberatore and Schaerf [1998] address arbitration in general, while Konieczny and Pino Pérez [2002] consider a general approach in which merging takes place with respect to a set of global constraints, or formulas that must hold in the merging. We examine these latter two approaches in detail below.

Konieczny et al. [2002] describe a very general framework in which a family of merging operators is parametrised by a distance between interpretations and aggregating functions. More or less concurrently, Meyer [2001] proposed a general approach to formulating merging functions based on ordinal conditional functions [Spohn 1988]. Booth [2002] also considers the problem of an agent merging information from different sources, via what is called *social contraction*. Last, much work has been carried out in merging possibilistic knowledge bases; we mention here, e.g., the method by Benferhat et al. [2003].

We next describe the approaches by Liberatore and Schaerf [1998] and by Konieczny and Pino Pérez [2002], since we use the intuitions underlying these approaches as the basis for our merging technique. First, Liberatore and Schaerf [1998] consider merging two belief bases built on the intuition that models of the merged bases should be taken from those of each belief base closest to the other. This is called an *arbitration operator* (Konieczny and Pino Pérez [2002] call it a *commutative revision operator*). They consider a propositional language over a finite set of atoms; consequently their merging operator can be expressed as a binary operator on formulas. The following postulates characterise this operator:

DEFINITION  4.   $\diamond$ *is an* arbitration operator *if $\diamond$ satisfies the following postulates.*

($LS1$).  $\alpha \diamond \beta \equiv \beta \diamond \alpha$.

($LS2$).  $\alpha \wedge \beta$ *implies* $\alpha \diamond \beta$.

($LS3$).  *If* $\alpha \wedge \beta$ *is satisfiable then* $\alpha \diamond \beta$ *implies* $\alpha \wedge \beta$.

($LS4$).  $\alpha \diamond \beta$ *is unsatisfiable iff* $\alpha$ *is unsatisfiable and* $\beta$ *is unsatisfiable.*

($LS5$).  *If* $\alpha_1 \equiv \alpha_2$ *and* $\beta_1 \equiv \beta_2$ *then* $\alpha_1 \diamond \beta_1 \equiv \alpha_2 \diamond \beta_2$.

($LS6$).  $\alpha \diamond (\beta_1 \vee \beta_2) = \begin{cases} \alpha \diamond \beta_1 & or \\ \alpha \diamond \beta_2 & or \\ (\alpha \diamond \beta_1) \vee (\alpha \diamond \beta_2). \end{cases}$

($LS7$).  $(\alpha \diamond \beta)$ *implies* $(\alpha \vee \beta)$.

($LS8$).  *If* $\alpha$ *is satisfiable then* $\alpha \wedge (\alpha \diamond \beta)$ *is satisfiable.*

The first postulate asserts that merging is commutative, while the next two assert that, for mutually consistent formulas, merging corresponds to their conjunction. ($LS5$) ensures that the operator is independent of syntax, while ($LS6$) provides a "factoring" postulate, analogous to a similar factoring result in (AGM-style) belief revision and contraction. Postulate ($LS7$) can be taken as distinguishing $\diamond$ from other such operators; it asserts that the result of merging implies the disjunction of the original formulas. The last postulate informally constrains the result of merging so that each operator "contributes to" (i.e., is consistent with) the final result.

Next, Konieczny and Pino Pérez [2002] consider the problem of merging possibly contradictory belief bases. To this end, they consider finite multisets of the form $\Psi = \{K_1, \ldots, K_n\}$. They assume that the belief sets $K_i$ are consistent and finitely representable, and so representable by a formula. $K^n$ is the multiset consisting of $n$ copies of $K$. Following Konieczny and Pino Pérez [2002], let $\Delta^\mu(\Psi)$ denote the result of merging the multiset $\Psi$ of belief bases given the entailment-based integrity constraint expressed by $\mu$. The intent is that $\Delta^\mu(\Psi)$ is the belief base closest to the belief multiset $\Psi$. They provide the following set of postulates (multiset union is denoted by $\cup$):

DEFINITION 5.  *Let* $\Psi$ *be a multiset of sets of formulas, and* $\phi$, $\mu$ *formulas* (*all possibly subscripted or primed*). *Then,* $\Delta$ *is an* IC *merging operator if it satisfies the following postulates.*

($IC0$).  $\Delta^\mu(\Psi) \vdash \mu$.

($IC1$).  *If* $\mu \nvdash \bot$ *then* $\Delta^\mu(\Psi) \nvdash \bot$.

($IC2$).  *If* $\bigwedge \Psi \nvdash \neg\mu$ *then* $\Delta^\mu(\Psi) \equiv \bigwedge \Psi \wedge \mu$.

($IC3$).  *If* $\Psi_1 \equiv \Psi_2$ *and* $\mu_1 \equiv \mu_2$ *then* $\Delta^{\mu_1}(\Psi_1) \equiv \Delta^{\mu_2}(\Psi_2)$.

($IC4$).  *If* $\phi \vdash \mu$ *and* $\phi' \vdash \mu$ *then* $\Delta^\mu(\phi \cup \phi') \wedge \phi \nvdash \bot$ *implies* $\Delta^\mu(\phi \cup \phi') \wedge \phi' \nvdash \bot$.

($IC5$).  $\Delta^\mu(\Psi_1) \wedge \Delta^\mu(\Psi_2) \vdash \Delta^\mu(\Psi_1 \cup \Psi_2)$.

($IC6$).  *If* $\Delta^\mu(\Psi_1) \wedge \Delta^\mu(\Psi_2) \nvdash \bot$ *then* $\Delta^\mu(\Psi_1 \cup \Psi_2) \vdash \Delta^\mu(\Psi_1) \wedge \Delta^\mu(\Psi_2)$.

($IC7$).  $\Delta^{\mu_1}(\Psi) \wedge \mu_2 \vdash \Delta^{\mu_1 \wedge \mu_2}(\Psi)$.

($IC8$).  *If* $\Delta^{\mu_1}(\Psi) \wedge \mu_2 \nvdash \bot$ *then* $\Delta^{\mu_1 \wedge \mu_2}(\Psi) \vdash \Delta^{\mu_1}(\Psi) \wedge \mu_2$.

($IC2$) states that, when consistent, the result of merging is simply the conjunction of the belief bases and integrity constraints. ($IC4$) asserts that when two belief bases disagree, merging does not give preference to one of them. ($IC5$) states that a model of two mergings is in the union of their merging. With ($IC5$) we get that if two mergings are consistent then their merging is implied by their conjunction. Note that merging operators are trivially commutative. $(IC7)$ and $(IC8)$ correspond to the extended AGM postulates $(K * 7)$ and $(K * 8)$ for revision (cf. [Alchourrón et al. 1985; Gärdenfors 1988]), but with respect to the integrity constraints.

## 2.3    Belief Change in Logic Programming

Most previous work on belief change for logic programs goes under the title of *update* [Foo and Zhang 1997; Przymusinski and Turner 1997; Zhang and Foo 1998; Alferes et al. 1998; 2000; Leite 2003; Inoue and Sakama 1999; Eiter et al. 2002; Zacarías et al. 2005; Delgrande et al. 2007]. Strictly speaking, however, such approaches generally do not address "update," at least insofar as the term is understood in the belief revision community. There, update refers to a belief change in response to a change in the world being modelled [Katsuno and Mendelzon 1992]; this notion of change is not taken into account in the above-cited work. A common feature of most update approaches is to consider a sequence $P_1$, $P_2, \ldots, P_n$ of programs where each $P_i$ is a logic program. For $P_i$, $P_j$, and $i > j$, the intuition is that $P_i$ has higher priority or precedence. Given such a sequence, a set of answer sets is determined that in some sense respects the ordering. This may be done by translating the sequence into a single logic program that contains an encoding of the priorities, or by treating the sequence as a prioritised logic program, or by some other appropriate method. The net result, one way or another, is to obtain a set of answer sets from such a program sequence, and not a single new program expressed in the language of the original logic programs. Hence, these approaches fall outside the general AGM belief revision paradigm.

However, various principles have been proposed for such approaches to logic program update. In particular, Eiter et al. [2002] consider the question of what principles the update of logic programs should satisfy. This is done by re-interpreting different AGM-style postulates for revising or updating classic knowledge bases, as well as introducing new principles. Among the latter, let us note the following:

*Initialisation.*  $\emptyset * P \equiv P$.

*Idempotency.*  $(P * P) \equiv P$.

*Tautology.*  If $Q$ is tautologous, then $P * Q \equiv P$.

*Absorption.*  If $Q = R$, then $((P * Q) * R) \equiv (P * Q)$.

*Augmentation.*  If $Q \subseteq R$, then $((P * Q) * R) \equiv (P * R)$.

In view of the failure of several of the discussed postulates in the approach of Eiter et al. [2002] (as well as in others), Osorio and Cuevas [2007] noted that for re-interpreting the standard AGM postulates in the context of logic programs, the logic underlying strong equivalence should be adopted. Since Osorio and Cuevas [2007] studied programs with strong negation, this led them to consider the logic $\mathbf{N}_2$, an extension of HT by allowing strong negation.[3] They also introduced a new principle, which they called *weak independence of syntax* (WIS), which they proposed any update operator should satisfy:

*WIS.*  If $Q \equiv_s R$, then $(P * Q) \equiv (P * R)$.

---

[3]$\mathbf{N}_2$ itself traces back to an extension of intuitionistic logic with strong negation, first studied by Nelson [1949].

Indeed, following this spirit, the above absorption and augmentation principles can be accordingly changed by replacing their antecedents by "$Q \equiv_s R$" and "$Q \models_s R$", respectively. We note that the WIS principle was also discussed in an update approach based on *abductive programs* [Zacarías et al. 2005].

Turning our attention to the few works on *revision* of logic programs, early work in this direction includes a series of investigations dealing with restoring consistency for programs possessing no answer sets (cf., e.g., Witteveen et al. [1994]). Other work uses logic programs under a variant of the stable semantics to specify database revision, i.e., the revision of knowledge bases given as sets of atomic facts [Marek and Truszczyński 1998]. Finally, an approach following the spirit of AGM revision is discussed by Kudo and Murai [2004]. In their work, they deal with the question of constructing revisions of form $P * A$, where $P$ is an extended logic program and $A$ is a conjunction of literals. They give a procedural algorithm to construct the revised programs; however no properties are analysed.

With respect to merging logic programs, we have already mentioned updating logic programs, which can also be considered as *prioritised logic program merging*. With respect to combining logic programs, Baral et al. [1991] describe an algorithm for combining a set of normal, stratified logic programs in which the union of the programs is also stratified. In their approach the combination is carried out so that a set of global integrity constraints, which is satisfied by individual programs, is also satisfied by the combination. Buccafurri and Gottlob [2002] present an interesting approach whereby rules in a given program encode desires for a corresponding agent. A predicate *okay* indicates that an atom is acceptable to an agent. Answer sets of these *compromise logic programs* represent acceptable compromises between agents. While it is shown that the joint fixpoints of such logic programs can be computed as answer sets, and complexity results are presented, the approach is not analysed from the standpoint of properties of merging. Sakama and Inoue [2008] address what they call the *generous* and *rigorous coordination* of logic programs in which, given a pair of programs $P_1$ and $P_2$, a program $Q$ is found whose answer sets are equal to the union of the answer sets of $P_1$ and $P_2$ in the first case, and their intersection in the second. As the authors note, this approach and its goals are distinct from program merging.

## 3.  BELIEF CHANGE IN ASP BASED ON SE MODELS

In AGM belief change, an agent's beliefs may be abstractly characterised in various different ways. In the classical AGM approach an agent's beliefs are given by a *belief set*, i.e., a deductively-closed set of sentences. As well, an agent's beliefs may also be characterised abstractly by a set of interpretations or *possible worlds*; these would correspond to models of the agent's beliefs. Last, as proposed in the Katsuno-Mendelzon formulation, and given the assumption of a finite language, an agent's beliefs can be specified by a formula. Given a finite language, it is straightforward to translate between these representations.

In ASP, there are notions analogous to the above for specifying an agent's beliefs. Though we do not get into it here, the notion of *strong equivalence* of logic programs can be employed to define a (*logic program*) *belief set*. Indeed, a set of well-defined SE models characterises a class of equivalent logic programs. Hence, the set of SE models of a program can be considered as the *proposition* expressed by the program. Continuing this analogy, a specific logic program can be considered to correspond to a formula or set of formulas in classical logic.

### 3.1   Expanding Logic Programs

Belief *expansion* is a belief change operator that is much more basic than revision or contraction, and in a certain sense is *prior* to revision and contraction (since in the AGM approach revision and contraction postulates make reference to expansion). Hence, it is of interest to examine expansion from the point of view of logic programs. As well, it proves to be the case that expansion in logic programs is of interest in its own right.

The next definition corresponds model-theoretically with the usual definition of expansion in AGM belief change.

DEFINITION 6. *For logic programs $P$ and $Q$, define the* expansion *of $P$ and $Q$, $P + Q$, to be a logic program $R$ such that $SE(R) = SE(P) \cap SE(Q)$.*

For illustration, consider the following examples:[4]

(1)  $\{p \leftarrow\} + \{\bot \leftarrow p\}$ has no SE models.

(2)  $\{p \leftarrow q\} + \{\bot \leftarrow p\}$ has SE model $(\emptyset, \emptyset)$.

(3)  $\{p \leftarrow\} + \{q \leftarrow p\} \equiv_s \{p \leftarrow\} + \{q \leftarrow\} \equiv_s \{p \leftarrow, q \leftarrow\}$.

(4)  $\{p \leftarrow \sim q\} + \{q \leftarrow \sim p\} \equiv_s \left\{ \begin{array}{l} p \leftarrow \sim q \\ q \leftarrow \sim p \end{array} \right\}$.

(5)  $\left\{ \begin{array}{l} p \leftarrow \sim q \\ q \leftarrow \sim p \end{array} \right\} + \{p \leftarrow q\} \equiv_s \left\{ \begin{array}{l} p \leftarrow q \\ p \leftarrow \sim q \end{array} \right\}$.

(6)  $\left\{ \begin{array}{l} p \leftarrow \sim q \\ q \leftarrow \sim p \end{array} \right\} + \{p; q \leftarrow\} \equiv_s \{p; q \leftarrow\}$.

(7)  $\{p; q \leftarrow\} + \{\bot \leftarrow q\} \equiv_s \left\{ \begin{array}{l} p \leftarrow \\ \bot \leftarrow q \end{array} \right\}$.

(8)  $\{p; q \leftarrow\} + \{\bot \leftarrow p, q\} \equiv_s \left\{ \begin{array}{l} p; q \leftarrow \\ \bot \leftarrow p, q \end{array} \right\}$.

Belief expansion has desirable properties. The following all follow straightforwardly from the definition of expansion with respect to SE models.

THEOREM 1. *Let $P$ and $Q$ be logic programs. Then:*

*(1)  $P + Q$ is a logic program.*

*(2)  $P + Q \models_s P$.*

*(3)  If $P \models_s Q$, then $P + Q \equiv_s P$.*

*(4)  If $P \models_s Q$, then $P + R \models_s Q + R$.*

*(5)  If $SE(P)$ and $SE(Q)$ are well-defined, then so is $SE(P + Q)$.*

*(6)  If $SE(P)$ and $SE(Q)$ are complete, then so is $SE(P + Q)$.*

*(7)  If $Q \equiv_s \emptyset$, then $P + Q \equiv_s P$.*

While these results are indeed elementary, following as they do from the monotonicity of the SE interpretations framework, they are still of interest. Notably, virtually every previous approach to updating logic programs has trouble with the last property, expressing a *tautology* postulate. Here, expansion by a tautologous program presents no problem, as it corresponds to an intersection with the set of all SE interpretations. We note also that the other principles mentioned earlier—*initialisation*, *idempotency*, *absorption*, and *augmentation*—are trivially satisfied by expansion.

---

[4]Unless otherwise noted, we assume that the language of discourse in each example consists of just the atoms mentioned.

In classical logic, the expansion of two formulas can be given in terms of the intersection of their models. It should be clear from the preceding that the appropriate notion of the set of "models" of a logic program is given by a set of SE models, and not by a set of answer sets. Hence, there is no natural notion of expansion that is given in terms of answer sets. For instance, in Example 3, we have $AS(\{p \leftarrow\}) = \{\{p\}\}$ and $AS(\{q \leftarrow p\}) = \{\emptyset\}$ while $AS(\{p \leftarrow, q \leftarrow p\}) = \{\{p, q\}\}$. Likewise, in Example 4, the intersection of $AS(\{\{p \leftarrow \sim q\}\}) = \{\{p\}\}$ and $AS(\{\{q \leftarrow \sim p\}\}) = \{\{q\}\}$ is empty, whereas $AS(\{p \leftarrow \sim q, \ q \leftarrow \sim p\}) = \{\{p\}, \{q\}\}$.

## 3.2 Revising Logic Programs

We next turn to specific operators for belief revision. As discussed earlier, for a revision $P * Q$, we suggest that the most natural distance-based notion of revision for logic programs uses set containment as the appropriate means of relating SE interpretations. Hence, we begin by considering set-containment based revision. Thus, $P * Q$ will be a logic program whose SE models are a subset of the SE models of $Q$, comprising just those models of $Q$ that are closest to those of $P$. Following the development of this operator we also consider cardinality-based revision, as a point of contrast. While these two approaches correspond to the two best-known ways of incorporating distance based revision, they are not exhaustive and any other reasonable notion of distance could also be employed.

### 3.2.1 *Set-Containment Based Revision.*

The following definition gives, for sets of interpretations $E_1$, $E_2$, the subset of $E_1$ that is closest to $E_2$, where the notion of "closest" is given in terms of symmetric difference.

DEFINITION 7. *Let $E_1$, $E_2$ be two sets of either classical or SE interpretations. Then:*

$$\sigma(E_1, E_2) = \{A \in E_1 \mid \exists B \in E_2 \ \text{such that}$$
$$\forall A' \in E_1, \forall B' \in E_2, A' \ominus B' \not\subset A \ominus B\}.$$

It might seem that we could now define the SE models of $P * Q$ to be given by $\sigma(SE(Q), SE(P))$. However, for our revision operator to be meaningful, it must also produce a *well-defined* set of SE models. Unfortunately, Definition 7 does not preserve well-definedness. For an example, consider $P = \{\bot \leftarrow p\}$ and $Q = \{p \leftarrow \sim p\}$. Then, $SE(P) = \{(\emptyset, \emptyset)\}$ and $SE(Q) = \{(\emptyset, p), (p, p)\}$, and so $\sigma(SE(Q), SE(P)) = \{(\emptyset, p)\}$. However $\{(\emptyset, p)\}$ is not well-defined.

The problem is that for programs $P$ and $Q$, there may be an SE model $(X, Y)$ of $Q$ with $X \subset Y$ such that $(X, Y) \in \sigma(SE(Q), SE(P))$ but $(Y, Y) \notin \sigma(SE(Q), SE(P))$. Hence, in defining $P * Q$ in terms of $\sigma(SE(Q), SE(P))$, we must elaborate the set $\sigma(SE(Q), SE(P))$ in some fashion to obtain a well-defined set of SE models.

In view of this, our approach is based on the following idea to obtain a well-defined set of models of $P * Q$ based on the notion of distance given in $\sigma$:

(1) Determine the "closest" models of $Q$ to $P$ of form $(Y, Y)$.

(2) Determine the "closest" models of $Q$ to $P$ limited to models $(X, Y)$ of $Q$ where $(Y, Y)$ was found in the first step.

Thus, we give preference to potential answer sets, in the form of models $(Y, Y)$, and then to general models.

DEFINITION 8. *For logic programs $P$ and $Q$, define the* revision *of $P$ by $Q$, $P * Q$, to be a logic program such that:*

$$\text{if } SE(P) = \emptyset, \text{ then } SE(P * Q) = SE(Q);$$

*otherwise*

$$SE(P * Q) = \{(X, Y) \mid Y \in \sigma(Mod(Q), Mod(P)), X \subseteq Y,$$
$$\text{and if } X \subset Y \text{ then } (X, Y) \in \sigma(SE(Q), SE(P))\}.$$

As is apparent, $SE(P * Q)$ is well-defined, and thus is representable through a canonical logic program. Furthermore, over classical models, the definition of revision reduces to that of containment-based revision in propositional logic [Satoh 1988]. As we show below, the result of revising $P$ by $Q$ is identical to that of expanding $P$ by $Q$ whenever $P$ and $Q$ possess common SE models. Hence, all previous examples of non-empty expansions are also valid program revisions. We have the following examples of revision that do not reduce to expansion.[5]

(1) $\{p \leftarrow \sim p\} * \{\bot \leftarrow p\} \equiv_s \{\bot \leftarrow p\}$.
    Over the language $\{p, q\}$, $\bot \leftarrow p$ has SE models $(\emptyset, \emptyset)$, $(\emptyset, q)$, and $(q, q)$.

(2) $\left\{ \begin{array}{l} p \leftarrow \\ q \leftarrow \end{array} \right\} * \{\bot \leftarrow q\} \equiv_s \left\{ \begin{array}{l} p \leftarrow \\ \bot \leftarrow q \end{array} \right\}$.
    The first program has a single SE model, $(pq, pq)$, while the second has three, $(\emptyset, \emptyset)$, $(\emptyset, p)$, and $(p, p)$. Among the latter, $(p, p)$ has the least pairwise symmetric difference to $(pq, pq)$. The program induced by the singleton set $\{(p, p)\}$ of SE models is

    $$\{p \leftarrow, \ \bot \leftarrow q\}.$$

(3) $\left\{ \begin{array}{l} p \leftarrow \\ q \leftarrow \end{array} \right\} * \{\bot \leftarrow p, q\} \equiv_s \left\{ \begin{array}{l} p; q \leftarrow \\ \bot \leftarrow p, q \end{array} \right\}$.
    Thus, if one originally believes that $p$ and $q$ are true, and revises by the fact that one is false, then the result is that precisely one of $p$, $q$ is true.

(4) $\left\{ \begin{array}{l} \bot \leftarrow \sim p \\ \bot \leftarrow \sim q \end{array} \right\} * \{\bot \leftarrow p, q\} \equiv_s \left\{ \begin{array}{l} \bot \leftarrow \sim p, \sim q \\ \bot \leftarrow p, q \end{array} \right\}$.
    Observe that the classical models in the programs here are exactly the same as above. This example shows that the use of SE models provides finer "granularity" compared to using classical models of programs together with known revision techniques.

(5) $\left\{ \begin{array}{l} \bot \leftarrow p \\ \bot \leftarrow q \end{array} \right\} * \{p; q \leftarrow\} \equiv_s \left\{ \begin{array}{l} p; q \leftarrow \\ \bot \leftarrow p, q \end{array} \right\}$.

We next rephrase the Katsuno-Mendelzon postulates for belief revision. Here, $*$ is a function from ordered pairs of logic programs to logic programs.

$(RA1)$. $P * Q \models_s Q$.

$(RA2)$. If $P + Q$ is satisfiable, then $P * Q \equiv_s P + Q$.

$(RA3)$. If $Q$ is satisfiable, then $P * Q$ is satisfiable.

---

[5]Note that $\{p \leftarrow \sim p\}$ has SE models but no answer sets.

$(RA4)$. If $P_1 \equiv_s P_2$ and $Q_1 \equiv_s Q_2$, then $P_1 * Q_1 \equiv_s P_2 * Q_2$.

$(RA5)$. $(P * Q) + R \models_s P * (Q + R)$.

$(RA6)$. If $(P * Q) + R$ is satisfiable, then $P * (Q + R) \models_s (P * Q) + R$.

We obtain that logic program revision as given in Definition 8 satisfies the first five of the revision postulates. Unsurprisingly, this is analogous to set-containment based revision in propositional logic.

THEOREM 2. *The logic program revision operator $*$ from Definition 8 satisfies postulates $(RA1)$-$(RA5)$.*

The fact that our revision operator does not satisfy $(RA6)$ can be seen by the following example: Consider

$$P = \{p; \sim p, \ q \leftarrow p, \ r \leftarrow p, \ s \leftarrow p, \ \bot \leftarrow \sim p, q,$$
$$\bot \leftarrow \sim p, r, \ \bot \leftarrow \sim p, s\},$$
$$Q = \{p; r, \ \bot \leftarrow q, \ \bot \leftarrow p, r, \ \bot \leftarrow p, s, \ s; \sim s \leftarrow r\},$$
$$R = \{p; r, \ \bot \leftarrow q, \ \bot \leftarrow p, r, \ \bot \leftarrow p, s, \ s \leftarrow r\}.$$

Straightforward computations show that

$$SE(P * (Q + R)) = \{(rs, rs), (p, p)\} \qquad \text{while}$$
$$SE((P * Q) + R) = \{(p, p)\}.$$

So, $P * (Q + R) \not\models_s (P * Q) + R$. Since $SE((P * Q) + R) \neq \emptyset$, this shows that $(RA6)$ indeed fails.

Last, we have the following result concerning other principles for updating logic programs listed earlier:

THEOREM 3. *Let $P$ and $Q$ be logic programs. Then, $P * Q$ satisfies initialisation, idempotency, tautology, and absorption with respect to strong equivalence.*

Augmentation, however, does not hold, nor would one expect it to hold in a distance-based approach. For example, consider the case where $P$, $Q$, and $R$ are characterised by models $SE(P) = \{(a, a), (ab, ab)\}$, $SE(Q) = \{(ab, ab), (ac, ac), (b, b)\}$, and $SE(R) = \{(ac, ac), (b, b)\}$. Thus $SE(R) \subseteq SE(Q)$. We obtain that $SE(P * Q) = SE(P + Q) = \{(ab, ab)\}$, and thus $SE((P*Q)*R) = \{(b, b)\}$. However, $SE(P*R) = \{(ac, ac), (b, b)\}$, contradicting augmentation.

Definition 8 seems to be the most natural possibility for constructing a set-containment based revision operator. However, it is not the only such possibility. We next briefly discuss an alternative definition for revision. The idea here is that for the revision of $P$ by $Q$, we select the closest models of $Q$ to $P$, and then add interpretations to make the result well-defined.

DEFINITION 9. *For logic programs $P$ and $Q$, define the* weak revision *of $P$ by $Q$ to be a logic program $P *_w Q$ such that:*

$$\text{if } SE(P) = \emptyset, \text{ then } SE(P *_w Q) = SE(Q);$$

*otherwise*

$$SE(P *_w Q) = \sigma(SE(Q), SE(P)) \cup$$
$$\{(Y, Y) \mid (X, Y) \in \sigma(SE(Q), SE(P)) \text{ for some } X\}.$$

The drawback to this approach is that it introduces possibly irrelevant interpretations in order to obtain well-definedness. As well, Definition 8 appears to be the more natural. Consider the following example, which also serves to distinguish Definition 8 from Definition 9. Let

$$P = \{\bot \leftarrow p, \; \bot \leftarrow q, \; \bot \leftarrow r\},$$
$$Q = \{\; r, \; p \leftarrow q, \; p \leftarrow \sim q \;\}.$$

Then, we get the following SE models:

$$SE(P) = \{(\emptyset, \emptyset)\},$$
$$SE(Q) = \{(r, pqr), (pr, pr), (pr, pqr), (pqr, pqr)\},$$

and

$$SE(P * Q) = \{(pr, pr)\},$$
$$SE(P *_w Q) = SE(Q) \setminus \{(pr, pqr)\}.$$

Consequently, $P * Q$ is given by the program $\{p, \; \bot \leftarrow q, r\}$. Thus, in this example, $P * Q$ gives the desired result, preserving the falsity of $q$ from $P$, while incorporating the truth of $r$ and $p$ from $Q$. This then reflects the assumption of minimal change to the program being revised, in this case $P$. $P *_w Q$ on the other hand represents a very cautious approach to program revision.

Finally, we have that our definition of revision is strictly stronger than the alternative given by $*_w$:

THEOREM 4. *Let $P$ and $Q$ be programs. Then, $P * Q \models_s P *_w Q$.*

For completeness, let us mention that enforcing well-definedness by simply determining the "closest" models of $Q$ to $P$ of form $(Y, Y)$ is inadequate. For our motivating example, we would obtain $SE(\{p \leftarrow \sim p\} * \{\bot \leftarrow p\}) = \emptyset$, violating the key postulate $(RA3)$, that the result of revising by a satisfiable program results in a satisfiable revision.

3.2.2 *Cardinality-Based Revision.* We next briefly recapitulate the previous development but in terms of cardinality-based revision. Define, for two sets $E_1$, $E_2$ of interpretations, the subset of $E_1$ that is closest to $E_2$, where the notion of "closest" is now given in terms of cardinality:

DEFINITION 10. *Let $E_1$, $E_2$ be two sets of either classical or SE interpretations. Then:*

$$\sigma_{||}(E_1, E_2) = \{A \in E_1 \mid \exists B \in E_2 \; \text{such that}$$
$$\forall A' \in E_1, \forall B' \in E_2, |A' \ominus B'| \not\prec |A \ominus B|\}.$$

As with set containment-based revision, we must ensure that our operator results in a well-defined set of SE models. Again, we first give preference to potential answer sets, in the form of models $(Y, Y)$, and then to general models.

DEFINITION 11. *For logic programs $P$ and $Q$, define the* (cardinality-based) *revision of $P$ by $Q$, $P *_c Q$, to be a logic program such that:*

$$\text{if } SE(P) = \emptyset, \text{ then } SE(P *_c Q) = SE(Q);$$

*otherwise*

$$SE(P *_c Q) = \{(X, Y) \mid Y \in \sigma_{||}(Mod(Q), Mod(P)), X \subseteq Y,$$
$$\textit{and if } X \subset Y \textit{ then } (X, Y) \in \sigma_{||}(SE(Q), SE(P))\}.$$

$P *_c Q$ can be seen to be well-defined, and so can be represented through a canonical logic program. As well, over classical, propositional models the definition reduces to cardinality-based revision in propositional logic [Dalal 1988].

We observe from the respective definitions that

$$SE(P *_c Q) \subseteq SE(P * Q).$$

That the two revision operators differ is easily shown: For example, if

$$P = \left\{ \begin{array}{l} p \leftarrow \\ q \leftarrow \\ r \leftarrow \end{array} \right\} \text{ and } Q = \left\{ \begin{array}{l} p \,; \, q \leftarrow \\ r \leftarrow q \\ \leftarrow p, r \end{array} \right\}$$

we get $SE(P) = \{(pqr, pqr)\}$ and $SE(Q) = \{(p, p), (qr, qr)\}$. This yields $SE(P * Q) = \{(p, p), (qr, qr)\}$ while $SE(P *_c Q) = \{(qr, qr)\}$.

It can be observed that $P *_c Q$ yields the same results as $P * Q$ for the five examples given in the previous subsection. However, cardinality-based revision fully aligns with the AGM postulates:

THEOREM 5. *Let $P$ and $Q$ be logic programs. Then, $P *_c Q$ satisfies postulates $(RA1)$ – $(RA6)$.*

As well, the following result is straightforward:

THEOREM 6. *Let $P$ and $Q$ be logic programs. Then, $P *_c Q$ satisfies initialisation, idempotency, tautology, and absorption with respect to strong equivalence.*

3.2.3 *Remarks.* Both of our proposed approaches to revising logic programs are based on a notion of distance between SE models. In the first, a partial preorder was induced between SE models, while in the second a total preorder resulted. We note that any definition of distance that results in a partial (resp., total) preorder among SE models could have been used, with the same technical results obtaining (but not, of course, the same examples). Hence, these approaches are exemplars of the two most common types of revision, expressed in terms of differences among truth values of atoms in models. As such, our specific approaches can be seen as natural generalisations of the approaches of Satoh [1988] and Dalal [1988].

We have suggested earlier that the approach based on set containment is the more natural or plausible approach, even though it does not satisfy all of the AGM postulates. This is because the cardinality-based approach may make somewhat arbitrary distinctions in arriving at a total preorder over SE interpretations. Recall the example we used to illustrate the difference between the approaches:

$$SE(P) = \{(pqr, pqr)\} \text{ and } SE(Q) = \{(p, p), (qr, qr)\},$$

yielding

$$SE(P * Q) = \{(p, p), (qr, qr)\} \text{ and } SE(P *_c Q) = \{(qr, qr)\}.$$

Given that we have no information concerning the ontological import of the atoms involved, it seems somewhat arbitrary to decide (in the case of $*_c$) that $qr$ should take priority over $p$. As an alternative argument, consider where for some large $n$ we have

$$SE(P) = \{(p_1 \ldots p_{2n}, p_1 \ldots p_{2n})\} \quad \text{and}$$
$$SE(Q) = \{(p_1 \ldots p_{n+1}, p_1 \ldots p_{n+1}), (p_1 \ldots p_n, p_1 \ldots p_n)\}.$$

So, in this example it is quite arbitrary to select (as the cardinality-based approach does) $(p_1 \ldots p_{n+1}, p_1 \ldots p_{n+1})$ over $(p_1 \ldots p_n, p_1 \ldots p_n)$.

Let us finally remark that another plausible definition of an ordering underlying cardinality-based revision would be the following:

$$|(X_1, X_2)| \leq' |(Y_1, Y_2)| \text{ iff } |X_1| \leq |Y_1| \text{ and } |X_2| \leq |Y_2|.$$

However, this ordering yields a partial preorder, and a revision operator based on this notion of distance would be very similar to $P * Q$; in particular the postulate $(RA6)$ would not be satisfied. Since this operator is of at best marginal interest, we do not explore it further.

### 3.3 Complexity of Revision

We now turn to the complexity of our approach to revision. First, we recapitulate the complexity classes relevant in what follows. As usual, for any complexity class $C$, by co-$C$ we understand the class of all problems which are complementary to the problems in $C$. Furthermore, for $C$ as before and complexity class $A$, the notation $C^A$ stands for the *relativised version* of $C$, consisting of all problems which can be decided by Turing machines of the same sort and time bound as in $C$, only that the machines have access to an oracle for problems in $A$.

Four complexity classes are relevant here, viz. NP, $\Sigma_2^P$, $\Pi_2^P$, and $\Theta_2^P$, which are defined thus:

—NP consists of all decision problems which can be solved with a nondeterministic Turing machine working in polynomial time;

—$\Sigma_2^P = \text{NP}^{\text{NP}}$;

—$\Pi_2^P = \text{co-}\Sigma_2^P$; and

—$\Theta_2^P$ is the class of all problems solvable on a deterministic Turing machine in polynomial time asking on input $x$ a total of $O(\log |x|)$ many oracle calls to NP (thus, $\Theta_2^P$ is also denoted by $\text{P}^{\text{NP}[\log n]}$).

Observe that NP, $\Sigma_2^P$, and $\Pi_2^P$ are part of the polynomial hierarchy, which is given by the following sequence of objects: the initial elements are

$$\Delta_0^P = \Sigma_0^P = \Pi_0^P = \text{P};$$

and, for $i > 0$,

$$\Delta_i^P = \text{P}^{\Sigma_{i-1}^P}; \qquad \Sigma_i^P = \text{NP}^{\Sigma_{i-1}^P}; \qquad \Pi_i^P = \text{co-NP}^{\Sigma_{i-1}^P}.$$

Here, P is the class of all problems solvable on a deterministic Turing machine in polynomial time. It holds that $\Sigma_1^P = \text{NP}$, $\Sigma_2^P = \text{NP}^{\text{NP}}$, and $\Pi_2^P = \text{co-NP}^{\text{NP}}$. A problem is said to be at the *$k$-th level* of the polynomial hierarchy iff it is in $\Delta_{k+1}^P$ and either $\Sigma_k^P$-hard or $\Pi_k^P$-hard.

We first consider the worst-case complexity of our approach to set-containment based revision. The standard decision problem for revision in classical logic is:

Given formulas $P$, $Q$, and $R$, does $P * Q$ entail $R$?

Eiter and Gottlob [1992] showed that approaches to classical propositional revision are $\Pi_2^P$-complete. The next result shows that this property carries over to our approach for program revision.

THEOREM 7. *Deciding whether $P * Q \models_s R$ holds, for given GLPs $P$, $Q$, and $R$, is $\Pi_2^P$-complete. Moreover, hardness holds already for $P$ being a set of facts, $Q$ being positive or normal, and $R$ being a single fact.*

Although we do not show it here, we mention that the same results holds for the cautious revision operator $*_w$ as well.

For cardinality-based revision, we obtain the following result, again mirroring a similar behaviour for the classical case and being a consequence of the construction given in Section 5:

THEOREM 8. *Deciding whether $P *_c Q \models_s R$ holds, for given GLPs $P$, $Q$, and $R$, is in $\Theta_2^P$.*

## 4. MERGING LOGIC PROGRAMS

We denote (generalised) logic programs by $P_1$, $P_2$, ..., reserving $P_0$ for a program representing global constraints, as described later. For logic programs $P_1$, $P_2$, we define $P_1 \sqcap P_2$ to be a program with SE models equal to $SE(P_1) \cap SE(P_2)$ and $P_1 \sqcup P_2$ to be a program with SE models equal to $SE(P_1) \cup SE(P_2)$. By a *belief profile*, $\Psi$, we understand a sequence[6] $\langle P_1, \ldots, P_n \rangle$ of (generalised) logic programs. For $\Psi = \langle P_1, \ldots, P_n \rangle$ we write $\sqcap \Psi$ for $P_1 \sqcap \cdots \sqcap P_n$. We write $\Psi_1 \circ \Psi_2$ for the (sequence) concatenation of belief profiles $\Psi_1$, $\Psi_2$; and for logic program $P_0$ and $\Psi = \langle P_1, \ldots, P_n \rangle$ we abuse notation by writing $\langle P_0, \Psi \rangle$ for $\langle P_0, P_1, \ldots, P_n \rangle$. A belief profile $\Psi$ is *satisfiable* just if each component logic program is satisfiable. The set of SE models of $\Psi$ is given by $SE(\Psi) = SE(P_1) \times \cdots \times SE(P_n)$. For $\overline{S} \in SE(\Psi)$ such that $\overline{S} = \langle S_1, \ldots, S_n \rangle$, we use $S_i$ to denote the $i^{th}$ component of $\overline{S}$. Thus, $S_i \in SE(P_i)$. Analogously, the set of classical propositional models of $\Psi$ is given by $Mod(\Psi) = Mod(P_1) \times \cdots \times Mod(P_n)$; also we use $X_i$ to denote the $i^{th}$ component of $\overline{X} \in Mod(\Psi)$.

### 4.1 Arbitration Merging

For the first approach to merging, called *arbitration*, we consider models of $\Psi$ and select those models in which, in a global sense, the constituent models vary minimally. The result of arbitration is a logic program made up of SE models from each of these minimally-varying tuples. Note that, in particular, if a set of programs is jointly consistent, then there are models of $\Psi$ in which all constituent SE models are the same. That is, the models that vary minimally are those $\overline{S} \in SE(\Psi)$ in which $S_i = S_j$ for every $1 \le i, j \le n$; and merging is the same as simply taking the union of the programs.

The first definition provides a notion of distance between models of $\Psi$, while the second then defines merging in terms of this distance.

---

[6]This departs from usual practise, where a belief profile is usually taken to be a multiset.

Table I.    Examples on Arbitration Merging.

| $P_1$ | $P_2$ | $SE(\nabla(\langle P_1, P_2\rangle))$ | $\nabla(\langle P_1, P_2\rangle)$ |
|-------|-------|----------------------------------------|-----------------------------------|
| $\{p \leftarrow\}$ | $\{q \leftarrow\}$ | $\{(pq, pq)\}$ | $\{p \leftarrow, q \leftarrow\}$ |
| $\{p \leftarrow\}$ | $\{\leftarrow p\}$ | $\{(p, p), (\emptyset, \emptyset)\}$ | $\{p; \sim p \leftarrow\}$ |
| $\{p \leftarrow \sim p\}$ | $\{\leftarrow p\}$ | $\{(\emptyset, p), (p, p), (\emptyset, \emptyset)\}$ | $\{\}$ |
| $\{p \leftarrow, q \leftarrow\}$ | $\{\leftarrow p, q\}$ | $\{(pq, pq), (p, p), (q, q)\}$ | $\{p; q \leftarrow, p; \sim p \leftarrow, q; \sim q \leftarrow\}$ |
| $\{\bot \leftarrow \sim p, \bot \leftarrow \sim q\}$ | $\{\leftarrow p, q\}$ | $\{S \in SE(\emptyset) \mid S \neq (\emptyset, \emptyset)\}$ | $\{\bot \leftarrow \sim p, \sim q\}$ |
| $\{\bot \leftarrow p, \bot \leftarrow q\}$ | $\{p; q \leftarrow\}$ | $\{(\emptyset, \emptyset), (p, p), (q, q)\}$ | $\{\leftarrow p, q; \sim p \leftarrow, q; \sim q \leftarrow\}$ |

DEFINITION 12.    *Let $\Psi = \langle P_1, \ldots, P_n\rangle$ be a satisfiable belief profile and let $\overline{S}, \overline{T}$ be two SE models of $\Psi$ (or two classical models of $\Psi$).*

*Then, define $\overline{S} \leq_a \overline{T}$, if $S_i \ominus S_j \subseteq T_i \ominus T_j$ for every $1 \leq i < j \leq n$.*

Clearly, $\leq_a$ is a partial preorder. In what follows, let $Min_a(N)$ denote the set of all minimal elements of a set $N$ of tuples relative to $\leq_a$, i.e.,

$$Min_a(N) = \{\overline{S} \in N \mid \overline{T} \leq_a \overline{S} \text{ implies } \overline{S} \leq_a \overline{T} \text{ for all } \overline{T} \in N\}.$$

Preparatory for our central definition to arbitration merging, we furthermore define, for a set $N$ of $n$-tuples,

$$\cup N = \{S_i \mid \text{ for } \overline{S} \in N, \overline{S} = \langle S_1, \ldots, S_n\rangle \text{ and } i \in \{1, \ldots, n\}\}.$$

DEFINITION 13.    *Let $\Psi = \langle P_1, \ldots, P_n\rangle$ be a belief profile. Then, the* arbitration merging, *or simply* arbitration, *of $\Psi$, is a logic program $\nabla(\Psi)$ such that*

$$SE(\nabla(\Psi)) = \{(X, Y) \mid Y \in \cup Min_a(Mod(\Psi)), X \subseteq Y,$$
$$\text{and if } X \subset Y \text{ then } (X, Y) \in \cup Min_a(SE(\Psi))\},$$

*providing $\Psi$ is satisfiable, otherwise, if $P_i$ is unsatisfiable for some $1 \leq i \leq n$, define $\nabla(\Psi) = \nabla(\langle P_1, \ldots, P_{i-1}, P_{i+1}, \ldots, P_n\rangle)$.*

For illustration, consider the belief profile

$$\langle P_1, P_2\rangle = \langle\{p \leftarrow, u \leftarrow\}, \{\leftarrow p, v \leftarrow\}\rangle. \tag{2}$$

Since $SE(P_1) = \{(pu, pu), (pu, puv), (puv, puv)\}$ and $SE(P_2) = \{(v, v), (v, uv), (uv, uv)\}$, we obtain nine SE models for $SE(\langle P_1, P_2\rangle)$. Among them, we find a unique $\leq_a$-minimal one, yielding $Min_a(SE(\langle P_1, P_2\rangle)) = \{\langle(puv, puv), (uv, uv)\rangle\}$. Similarly, $\langle P_1, P_2\rangle$ has a single $\leq_a$-minimal collection of pairs of classical models, viz. $Min_a(Mod(\langle P_1, P_2\rangle)) = \{\langle puv, uv\rangle\}$. Accordingly, we get

$$\cup Min_a(Mod(\langle P_1, P_2\rangle)) = \{puv, uv\},$$
$$\cup Min_a(SE(\langle P_1, P_2\rangle)) = \{(puv, puv), (uv, uv)\}, \text{ and}$$
$$SE(\nabla(\langle P_1, P_2\rangle)) = \cup Min_a(SE(\langle P_1, P_2\rangle)).$$

We thus obtain the program $\nabla(\langle P_1, P_2\rangle) = \{p; \sim p \leftarrow, u \leftarrow, v \leftarrow\}$ as the resultant arbitration of $P_1$ and $P_2$.

For further illustration, consider the technical examples given in Table I.

We note that merging normal programs often leads to disjunctive or generalised programs. Although plausible, this is also unavoidable because merging does not preserve the model intersection property of the reduced program satisfied by normal programs.

Moreover, we have the following general result.

THEOREM 9. *Let* $\Psi = \langle P_1, P_2 \rangle$ *be a belief profile, and define* $P_1 \diamond P_2 = \nabla(\Psi)$. *Then,* $\diamond$ *satisfies the following versions of the postulates of Definition 4.*

(LS1′). $P_1 \diamond P_2 \equiv_s P_2 \diamond P_1$.

(LS2′). $P_1 \sqcap P_2 \models_s P_1 \diamond P_2$.

(LS3′). *If* $P_1 \sqcap P_2$ *is satisfiable then* $P_1 \diamond P_2 \models_s P_1 \sqcap P_2$.

(LS4′). $P_1 \diamond P_2$ *is satisfiable iff* $P_1$ *is satisfiable and* $P_2$ *is satisfiable.*

(LS5′). *If* $P_1 \equiv_s P_2$ *and* $P_1' \equiv_s P_2'$ *then* $P_1 \diamond P_2 \equiv_s P_1' \diamond P_2'$.

(LS7′). $P_1 \diamond P_2 \models_s P_1 \sqcup P_2$.

(LS8′). *If* $P_1$ *and* $P_2$ *are satisfiable then* $P_1 \sqcap (P_1 \diamond P_2)$ *is satisfiable.*

## 4.2 Basic Merging

For the second approach to merging, programs $P_1, \ldots, P_n$ are merged with a target logic program $P_0$ so that the SE models in the merging will be drawn from models of $P_0$. This operator will be referred to as the (*basic*) *merging* of $P_1, \ldots, P_n$ with respect to $P_0$. The information in $P_0$ *must* hold in the merging, and so can be taken as *necessarily* holding. Konieczny and Pino Pérez [2002] call $P_0$ a set of *integrity constraints*, though this usage of the term differs from its usage in logic programs. Note that in the case where $SE(P_0)$ is the set of all SE models, the two approaches (of this section and the previous section) do not coincide, and that merging is generally a weaker operator than arbitration.

DEFINITION 14. *Let* $\Psi = \langle P_0, \ldots, P_n \rangle$ *be a belief profile and let* $\overline{S}, \overline{T}$ *be two SE models of* $\Psi$ *(or two classical models of* $\Psi$*).*

*Then, define* $\overline{S} \leq_b \overline{T}$, *if* $S_0 \ominus S_j \subseteq T_0 \ominus T_j$ *for every* $1 \leq j \leq n$.

As in the case of arbitration merging, $\leq_b$ is a partial preorder. Accordingly, let $Min_b(N)$ be the set of all minimal elements of a set $N$ of tuples relative to $\leq_b$. In extending our notation for referring to components of tuples, we furthermore define $N_0 = \{S_0 \mid \overline{S} \in N\}$. We thus can state our definition for basic merging as follows:

DEFINITION 15. *Let* $\Psi = \langle P_1, \ldots, P_n \rangle$ *be a belief profile. Then, the* basic merging, *or simply* merging, *of* $\Psi$, *is a logic program* $\Delta(\Psi)$ *such that*

$$SE(\Delta(\Psi)) = \{(X, Y) \mid Y \in Min_b(Mod(\Psi))_0, X \subseteq Y,$$
$$\text{and if } X \subset Y \text{ then } (X, Y) \in Min_b(SE(\Psi))_0\},$$

*providing* $\Psi$ *is satisfiable, otherwise, if* $P_i$ *is unsatisfiable for some* $1 \leq i \leq n$, *define* $\Delta(\Psi) = \Delta(\langle P_0, \ldots, P_{i-1}, P_{i+1}, \ldots, P_n \rangle)$.

Let us reconsider Programs $P_1$ and $P_2$ from (2) in the context of basic merging. To this end, we consider the belief profile $\langle \emptyset, \{p \leftarrow , u \leftarrow\}, \{\leftarrow p , v \leftarrow\} \rangle$. We are now faced with 27 SE models for $SE(\langle \emptyset, P_1, P_2 \rangle)$. Among them, we get the following $\leq_b$-minimal SE models

$$Min_b(SE(\langle \emptyset, P_1, P_2 \rangle)) = \{\langle(uv, uv), (puv, puv), (uv, uv)\rangle,$$
$$\langle(uv, puv), (puv, puv), (uv, uv)\rangle, \langle(puv, puv), (puv, puv), (uv, uv)\rangle\}$$

Table II. Examples on Basic Merging.

| $P_1$ | $P_2$ | $SE(\Delta(\langle \emptyset, P_1, P_2 \rangle))$ |
|---|---|---|
| $\{p \leftarrow\}$ | $\{q \leftarrow\}$ | $\{(pq, pq)\}$ |
| $\{p \leftarrow\}$ | $\{\leftarrow p\}$ | $\{(p, p), (\emptyset, \emptyset)\} \cup \{(p, \emptyset)\}$ |
| $\{p \leftarrow \sim p\}$ | $\{\leftarrow p\}$ | $\{(\emptyset, p), (p, p), (\emptyset, \emptyset)\}$ |
| $\{p \leftarrow \, , q \leftarrow\}$ | $\{\leftarrow p, q\}$ | $\{(pq, pq), (p, p), (q, q)\} \cup \{(p, pq), (q, pq)\}$ |
| $\{\perp \leftarrow \sim p \, , \perp \leftarrow \sim q\}$ | $\{\leftarrow p, q\}$ | $\{S \in SE(\emptyset) \mid S \neq (\emptyset, \emptyset)\}$ |
| $\{\perp \leftarrow p \, , \perp \leftarrow q\}$ | $\{p; q \leftarrow\}$ | $\{(\emptyset, \emptyset), (p, p), (q, q)\} \cup \{(p, \emptyset), (q, \emptyset)\}$ |

along with $Min_b(Mod(\langle \emptyset, P_1, P_2 \rangle)) = \{\langle uv, puv, uv \rangle, \langle puv, puv, uv \rangle\}$. We get:

$$Min_b(Mod(\langle \emptyset, P_1, P_2 \rangle))_0 = \{puv, uv\},$$
$$Min_b(SE(\langle \emptyset, P_1, P_2 \rangle))_0 = \{(uv, uv), (uv, puv), (puv, puv)\}, \text{ and}$$
$$SE(\Delta(\langle \emptyset, P_1, P_2 \rangle)) = Min_b(SE(\langle \emptyset, P_1, P_2 \rangle))_0 \,.$$

While arbitration resulted in $\nabla(\langle P_1, P_2 \rangle) = \{p; \sim p \leftarrow \, , u \leftarrow \, , v \leftarrow\}$, the more conservative approach of basic merging yields $\Delta(\langle \emptyset, P_1, P_2 \rangle) = \{u \leftarrow \, , v \leftarrow\}$.

We have just seen that basic merging adds "intermediate" SE models, viz. $(uv, puv)$, to the ones obtained in arbitration merging. This can also be observed on the examples given in Table I, where every second merging is weakened by the addition of such intermediate SE models. This is made precise in Theorem 11 below. We summarise the results in Table II but omit the programs $\Delta(\langle \emptyset, P_1, P_2 \rangle)$ because they are obtained from $\nabla(\langle P_1, P_2 \rangle)$ in Table I by simply dropping all rules of form $p; \sim p \leftarrow$ and $q; \sim q \leftarrow$, respectively.

The next example further illustrates the difference between arbitration and basic merging. Take $P_1 = \{p \leftarrow \, , q \leftarrow\}$ and $P_2 = \{\sim p \leftarrow \, , \sim q \leftarrow\}$. Then, we have that $SE(\nabla(\langle P_1, P_2 \rangle)) = \{(pq, pq), (\emptyset, \emptyset)\}$ and $SE(\Delta(\langle \emptyset, P_1, P_2 \rangle)) = SE(\emptyset)$. That is, in terms of programs, we obtain

$$\nabla(\langle P_1, P_2 \rangle) = \{p; \sim p \leftarrow, q; \sim q \leftarrow, \leftarrow p, \sim q, \leftarrow \sim p, q\} \text{ and } \Delta(\langle \emptyset, P_1, P_2 \rangle) = \emptyset \,.$$

THEOREM 10. *Let $\Psi$ be a belief profile, $P_0$ a program representing global constraints, and $\Delta$ as given in Definition 15. Then, $\Delta$ satisfies the following versions of the postulates of Definition 5:*

$(IC0')$. $\Delta(\langle P_0, \Psi \rangle) \models_s P_0$.

$(IC1')$. *If $P_0$ and $\Psi$ are satisfiable then $\Delta(\langle P_0, \Psi \rangle)$ is satisfiable.*

$(IC2')$. *If $\sqcap(P_0, \Psi)$ is satisfiable then $\Delta(\langle P_0, \Psi \rangle) \equiv_s P_0 \sqcap (\sqcap(\Psi))$.*

$(IC3')$. *If $P_0 \equiv_s P_0'$ and $\Psi \equiv_s \Psi'$ then $\Delta(\langle P_0, \Psi \rangle) \equiv_s \Delta(\langle P_0', \Psi' \rangle)$.*

$(IC4')$. *If $P_1 \models_s P_0$ and $P_2 \models_s P_0$ then:*
*if $\Delta(\langle P_0, P_1, P_2 \rangle) \sqcap P_1$ is satisfiable, then $\Delta(\langle P_0, P_1, P_2 \rangle) \sqcap P_2$ is satisfiable.*

$(IC5')$. $\Delta(\langle P_0, \Psi \rangle) \sqcap \Delta(\langle P_0, \Psi' \rangle) \models_s \Delta(\langle P_0, \Psi \circ \Psi' \rangle)$.

$(IC7')$. $\Delta(\langle P_0, \Psi \rangle) \sqcap P_1 \models_s \Delta(\langle P_0 \sqcap P_1, \Psi \rangle)$.

$(IC9')$. *Let $\Psi'$ be a permutation of $\Psi$. Then, $\Delta(\langle P_0, \Psi \rangle) \equiv_s \Delta(\langle P_0, \Psi' \rangle)$.*

We also obtain that arbitration merging is stronger than (basic) merging in the case of tautologous constraints in $P_0$.

THEOREM 11. *Let $\Psi$ be a belief profile. Then $\nabla(\Psi) \models_s \Delta(\langle \emptyset, \Psi \rangle)$.*

As well, for belief profile $\Psi = \langle P_1, P_2 \rangle$, we can express our merging operators in terms of the revision operator defined in Section 3.2.

THEOREM 12. *Let $\langle P_1, P_2 \rangle$ be a belief profile.*

*(1)* $\nabla(\langle P_1, P_2 \rangle) = (P_1 * P_2) \sqcup (P_2 * P_1)$.

*(2)* $\Delta(\langle P_1, P_2 \rangle) = P_2 * P_1$.

Note that in the second part of the preceding result, $P_1$ is regarded as a set of constraints (usually with name $P_0$) according to our convention for basic merging.

### 4.3 Complexity Analysis

In the previous section, the following decision problem was studied with respect to the revision operator $*$: Given GLPs $P$, $Q$, $R$, does $P * Q \models_s R$ hold? This problem was shown to be $\Pi_2^P$-complete. Accordingly, we give here results for the following problems:

(1) Given a belief profile $\Psi$ and a program $R$, does $\nabla(\Psi) \models_s R$ hold?

(2) Given a belief profile $\Psi$ and a program $R$, does $\Delta(\Psi) \models_s R$ hold?

By Theorem 12, it can be shown that the hardness result for the revision problem also applies to the respective problems in terms of merging. On the other hand, $\Pi_2^P$-membership can be obtained by a slight extension of the encodings given in the next section such that these extensions possess an answer set iff the respective problem (1) or (2) does *not* hold. Since checking whether a program has at least one answer set is a problem on the second of layer of the polynomial hierarchy, and our (extended) encodings are polynomial in the size of the encoded problems, the desired membership results follow.

THEOREM 13. *Given a belief profile $\Psi$ and a program $R$, deciding $\nabla(\Psi) \models_s R$ (resp., $\Delta(\Psi) \models_s R$) is $\Pi_2^P$-complete.*

## 5. COMPUTING BELIEF CHANGE VIA ANSWER SET PROGRAMMING

In this section, we provide encodings for our belief change operators in terms of fixed non-ground ASP programs. We recall that non-ground programs are defined over predicates of arbitrary arity which have either variables or constants as arguments. Such non-ground programs can be seen as a compact representation of large programs without variables (and thus as propositional programs), by considering the *grounding* of a program (recall that the grounding of a program $P$ is given by the union of the groundings of its rules, and the grounding of a rule $r \in P$ is the set obtained by all possible substitutions of variables in $r$ by constants occurring in $P$; cf. Dantsin et al. [2001] for a more thorough exposition). The non-ground programs we define in this section can be seen as queries which take the (propositional) programs subject to revision or merging as an input database. Thus, we follow here the tradition of meta-programming (see, e.g., the works of Delgrande et al. [2003], Eiter et al. [2003], and Gebser et al. [2008]).

Our encodings are given via certain language fragments of non-ground ASP such that their respective data complexity matches the complexity of the encoded task. Recall that data complexity addresses problems over programs $P \cup D$ where a non-ground program $P$ is fixed, while the input database $D$ (a set of ground atoms) is the input of the decision problem. As is well known, the data complexity of the problem whether $a$ is contained in all answer sets of $P \cup D$ is $\Pi_2^P$-complete for disjunctive programs (without weak constraints) [Dantsin et al. 2001] and $\Theta_2^P$-complete for normal programs with certain optimisation constructs (for instance, weak constraints over a single level; we introduce weak constraints later) [Buccafurri et al. 2000]. We use weak constraints for the encoding of the cardinality-based revision operator (although similar optimisation constructs could be used likewise) while for the set-based revision operator and the merging operators we require disjunctive programs. However, instead of coming up with encodings for the decisions problems discussed in the previous section, our ultimate goal here is to provide programs such that their answer sets characterise the SE models of the result of the encoded revision or merging problem.

Before we start with the ASP encodings, we have to fix how the programs subject to revision and merging are represented in our approach. To have a uniform setting in what follows, we use belief profiles $\Psi = \langle P_\alpha, \ldots, P_n \rangle$ where, for revision problems we have $\alpha = 1$ and $n = 2$ (and so $\langle P_1, P_2 \rangle$ here represents revision problem $P_1 * P_2$), for arbitration problems $\alpha = 1$ and $n \geq 2$, and for basic merging, we use $\alpha = 0$ and $n \geq 2$.

Given such a belief profile $\Psi = \langle P_\alpha, \ldots, P_n \rangle$, we use four ternary predicates, $phead$, $nhead$, $pbody$, and $nbody$ to represent $\Psi$. For each predicate, the first argument $i$ indices the program (i.e., $i$ is a number between $\alpha$ and $n$), the second argument contains the rule identifier $\#r$ of a rule $r \in P_i$, and the third argument is an atom, indicating that this atom occurs in the positive or negative head or the positive or negative body of rule $r \in P_i$, respectively. For example, let $\Psi = \langle P_1, P_2 \rangle$ with $P_1 = \{\leftarrow \sim p, \leftarrow \sim q\}$ and $P_2 = \{p; q \leftarrow, \leftarrow p, q\}$. We obtain the *relational representation of* $\Psi$ by[7]

$$[\Psi] = \{nbody(1,1,p), \ nbody(1,2,q),$$
$$phead(2,1,p), \ phead(2,1,q), \ pbody(2,2,p), \ pbody(2,2,q)\}.$$

Here, we just use numbers as rule identifiers, i.e. $\#(\leftarrow \sim p) = \#(p; q \leftarrow) = 1$ and $\#(\leftarrow \sim q) = \#(\leftarrow p, q) = 2$. The only necessary requirement is that different rules are assigned to different identifiers, i.e., $r \neq r'$ implies $\#r \neq \#r'$.

In general, we define the relational representation of a belief profile as follows.

DEFINITION 16. *Let* $\Psi = \langle P_\alpha, \ldots, P_n \rangle$ *be a belief profile. Then, the* relational representation *of* $\Psi$ *is given by*

$$[\Psi] = \bigcup_{i=\alpha}^{n} \bigcup_{r \in P_i} \Big( \{phead(i, \#r, a) \mid a \in H(r)^+\} \cup \{nhead(i, \#r, a) \mid a \in H(r)^-\} \cup$$
$$\{pbody(i, \#r, a) \mid a \in B(r)^+\} \cup \{nbody(i, \#r, a) \mid a \in B(r)^-\} \Big).$$

We assume here that all $i$ and $\#r$ are given as numbers. Note that some ASP solvers then require to define the domain of integers (for instance, via `#maxint` in the solver `DLV` [Leone et al. 2006]). Following datalog notation, we write, for a program $\Pi$ and a belief profile $\Psi$, $\Pi[\Psi]$ instead of $\Pi \cup [\Psi]$.

We provide our encodings in a modular way. That is, we introduce various sets of rules which implement different aspects required to solve the respective problem. We start with some basic modules, which are used in most of the encodings. Then, we provide our results for revision and conclude with the encodings for merging.

---

[7]Since we have here rules which are all simple facts, we omit the "$\leftarrow$"-symbol for rules.

## 5.1 Basic Modules

We start with a simple fragment which contains some domain predicates and fixes some designated identifiers.

DEFINITION 17.

$$
\begin{aligned}
\pi_{domain} \ = \ \{ & prog\_rule(P, R) \leftarrow \alpha(P, R, A), \ dom(A) \leftarrow \alpha(P, R, A) \ | \\
& \alpha \in \{ phead, pbody, nhead, nbody \} \} \cup \\
\{ & prog(P) \leftarrow prog\_rule(P, R), \\
& model(c) \leftarrow, \ model(t) \leftarrow, \ model(h) \leftarrow, \\
& prog\_model(c) \leftarrow, \ prog\_model(t) \leftarrow, \ prog\_model(h) \leftarrow \}.
\end{aligned}
$$

Predicates $prog\_rule(\cdot, \cdot)$, $dom(\cdot)$, and $prog(\cdot)$ are just used to gather some information from a conjoined input $[\Psi]$; the designated constants $c$, $t$, $h$ are used later on to distinguish between different guesses for models. In fact, $c$ refers to classical models while $h$ and $t$ refer to the first and second part of SE models, respectively.

The following code guesses such models for each program $P$ in the belief profile $\Psi$. The guess is accomplished in Rules (3) and (4) below which assign each atom $A$ in the domain to be $in(\cdot)$ or $out(\cdot)$.

DEFINITION 18.

$$
\begin{aligned}
\pi_{models} \ = \ \{ & in(P, A, M) \leftarrow \sim out(P, A, M), prog(P), dom(A), model(M), && (3) \\
& out(P, A, M) \leftarrow \sim in(P, A, M), prog(P), dom(A), model(M), && (4) \\
& \leftarrow in(P, A, h), out(P, A, t), && (5) \\
& diff(P, Q, A, M) \leftarrow in(P, A, M), out(Q, A, M), && (6) \\
& diff(P, Q, A, M) \leftarrow out(P, A, M), in(Q, A, M), && (7) \\
& same(P, Q, A, M) \leftarrow in(P, A, M), in(Q, A, M), && (8) \\
& same(P, Q, A, M) \leftarrow out(P, A, M), out(Q, A, M), && (9) \\
& ok(P, R, M) \leftarrow in(P, A, M), phead(P, R, A), model(M), && (10) \\
& ok(P, R, M) \leftarrow out(P, A, M), pbody(P, R, A), model(M), && (11) \\
& ok(P, R, M) \leftarrow in(P, A, M), nbody(P, R, A), prog\_model(M), && (12) \\
& ok(P, R, M) \leftarrow out(P, A, M), nhead(P, R, A), prog\_model(M), && (13) \\
& ok(P, R, h) \leftarrow in(P, A, t), nbody(P, R, A), && (14) \\
& ok(P, R, h) \leftarrow out(P, A, t), nhead(P, R, A), && (15) \\
& \leftarrow \sim ok(P, R, M), prog\_rule(P, R), model(M) \}. && (16)
\end{aligned}
$$

This allows us to draw a one-to-one correspondence between answer sets $S$ and models (resp., SE models) of the programs $P_i$ in the belief profile. Note that Rule (5) excludes such guesses where the corresponding SE model $(X, Y)$ would not satisfy $X \subseteq Y$. To make this intuition a bit more precise, let us define the following operator for a set $S$ of ground atoms and a number $i$:

$$
\begin{aligned}
\Pi^i_{Mod}(S) \ &= \ \{ a \mid in(i, a, c) \in S \}; \\
\Pi^i_{SE}(S) \ &= \ \big( \{ a \mid in(i, a, h) \in S \}, \{ b \mid in(i, b, t) \in S \} \big).
\end{aligned}
$$

The next Rules (6) – (9) indicate whether atom $A$ is assigned differently (via predicate $diff(\cdot, \cdot, \cdot, \cdot)$) or equally (via predicate $same(\cdot, \cdot, \cdot, \cdot)$) for two programs. These predicates are useful later.

The remaining Rules (10) – (15) tell us which rules (in which programs) are satisfied by the respective guess. Observe that (10) and (11) are applied to any forms of models (i.e., $h$, $t$, and $c$) while (12) and (13) are only applied to $t$ and $c$. Rules (14) and (15) finally take care of the fact that the first argument of an SE model has to be a model of the reduct. Therefore, we check whether the model given by the $t$-guess already eliminates rules. Note that such rules are satisfied by the $h$-guess in a trivial way. The last Rule (16) finally ensures that all rules of all programs are satisfied by our guesses.

We observe that the answer sets of the program $P[\Psi]$ where $P = \pi_{domain} \cup \pi_{models}$ are in a one-to-one correspondence to the models and SE models of belief profile $\Psi$. Formally we have that, given $\Psi = \langle P_\alpha, \ldots, P_n \rangle$,

$$\{(M, N) \mid M \in Mod(\Psi), N \in SE(\Psi)\} =$$
$$\{(\langle \Pi^\alpha_{Mod}(S), \ldots, \Pi^n_{Mod}(S) \rangle, \langle \Pi^\alpha_{SE}(S), \ldots, \Pi^n_{SE}(S) \rangle) \mid S \in AS(P[\Psi])\}.$$

We define one further module, which just orders the domain elements (i.e., the atoms in the given belief profile) using the standard ordering $<$ provided by the employed ASP solver. In particular, we define infimum, supremum, and successor with respect to this order. This is a standard technique used quite frequently in ASP; we require this later in order to "loop" over the domain elements. We also note that we could add such rules also for the program indices, as well as for the rule identifiers. However, since we assume them to be given by successive numbers, we omit such rules here. In fact, we use $N = M + 1$ instead of a designated successor predicate for program indices and rule identifiers. We add only rules which provide the minimal number $\alpha$ and resp. the maximal number $n$ of programs in the given profile $\Psi = \langle P_\alpha, \ldots, P_n \rangle$. The module thus looks as follows:

DEFINITION 19.

$$\begin{aligned}
\pi_{order} = \{ <(X, Y) \leftarrow dom(X), dom(Y), X < Y, \\
&nsucc(X, Z) \leftarrow lt(X, Y), lt(Y, Z), \\
&succ(X, Y) \leftarrow lt(X, Y), {\sim}nsucc(X, Y), \\
&ninf(X) \leftarrow lt(Y, X), \\
&nsup(X) \leftarrow lt(X, Y), \\
&inf(X) \leftarrow {\sim}ninf(X), dom(X), \\
&sup(X) \leftarrow {\sim}nsup(X), dom(X), \\
&minprog(P) \leftarrow prog(P), P = Q + 1, {\sim}prog(Q), \\
&maxprog(P) \leftarrow prog(P), Q = P + 1, {\sim}prog(Q)\}.
\end{aligned}$$

## 5.2 Encodings for Revision

5.2.1 *Cardinality-based Revision.* We are now prepared to encode the cardinality-based revision, following Definition 11. For our forthcoming encoding, we require optimisation constructs which are available in most ASP solvers. We shall use here the concept of *weak constraints* [Buccafurri et al. 2000] as used by DLV. A weak constraint (without weights and levels) is a rule of the form

$$\Leftarrow c_1, \ldots, c_m, \sim d_{m+1}, \ldots, \sim d_p. \tag{17}$$

The semantics of a program $P$ with such weak constraints is as follows: First, compute the answer sets of the program $P'$ given as $P$ without the weak constraints. Then, count for each answer set $S$ of $P$ the number of the grounded weak constraints which apply to $S$. By "apply to $S$", we mean that given a ground weak constraint of the form (17) $\{c_1, \ldots, c_m\} \subseteq S$ and $\{d_{m+1}, \ldots, d_p\} \cap S$ jointly hold. The answer sets of $P$ are then given by those answer sets $P'$ which have a minimal number of applied weak constraints.

Recall that for representing revision problems, we use belief profiles of the form $\langle P_1, P_2 \rangle$. We already have computed all models and SE models of both programs. We now have to (i) find those pairs $(M_1, M_2)$ of models $M_1 \in Mod(P_1)$, $M_2 \in Mod(P_2)$ which have a minimum number of differences, and (ii) find those pairs $(S_1, S_2)$ of SE models $S_1 \in Mod(S_1)$, $S_2 \in Mod(P_2)$ which are minimal with respect to the order $<$ defined in Definition 3. After isolating the respective answer sets we then can find the SE models of $P_1 *_c P_2$ (following Definition 11) quite straightforwardly. Indeed, Task (i) can easily be achieved by making use of weak constraints which penalise each mismatch, and a similar method is possible for Task (ii). However, Definition 3 requires a certain two-phased comparison of SE models. We thus count a mismatch in the $h$-models as many times as we have differences in the $t$- and $c$-models, respectively. Finally, if there was no difference in the $t$- or $c$-models, we still have to count the number of mismatches in the $h$-models. This is done by the third rule in the following program.

DEFINITION 20.

$$\begin{aligned}
\pi_{card} \; = \; \{ &\Leftarrow \mathit{diff}(1, 2, A, M), \mathit{prog\_model}(M), \\
&\Leftarrow \mathit{diff}(1, 2, A, h), \mathit{diff}(1, 2, B, M), \\
&\Leftarrow \mathit{diff}(1, 2, A, h), \sim\mathit{diff}(1, 2, B, M), \mathit{dom}(B), \mathit{prog\_model}(M), \\
&\mathit{selector}(2) \leftarrow \}.
\end{aligned}$$

We finally put things together. To this end, we first define a module which takes the models and SE models, respectively, of some selected program (in the case of revision, it is program $P_2$, thus $\mathit{selector}(2)$ is specified in $\pi_{card}$) and copies them into a designated predicate.

DEFINITION 21.

$$\begin{aligned}
\pi_{result} \; = \; \{ &total \leftarrow \sim nontotal, \\
&nontotal \leftarrow \sim total, \\
&\leftarrow nontotal, selector(S), in(S, A, t), out(S, A, c), \\
&\leftarrow nontotal, selector(S), out(S, A, t), in(S, A, c), \\
&resultH(A) \leftarrow selector(S), in(S, A, h), nontotal, \\
&resultH(A) \leftarrow selector(S), in(S, A, c), total, \\
&resultT(A) \leftarrow selector(S), in(S, A, c) \}.
\end{aligned}$$

The intuition for the module is as follows: we either generate a total SE model $(Y, Y)$ or a non-total SE model $(X, Y)$ with $X \subseteq Y$. Thus, the guess between predicates $total$ and $nontotal$. In case we want to derive a non-total SE model $(X, Y)$, we have to make sure that $Y$ coincides with the classical model we guessed.[8] This is done by the two constraints. The remaining lines fill the predicates $resultH$ and $resultT$ accordingly, where atoms in $resultH$ yield the $X$ of the SE model and atoms in $resultT$ yield the $Y$ of the SE model.

We formulate our first main result.

THEOREM 14. *For any set $S$ of non-ground atoms, let*

$$\rho(S) = (\{a \mid resultH(a) \in S\}, \{b \mid resultT(b) \in S\}).$$

*Moreover, let*

$$P_{card} = \pi_{domain} \cup \pi_{models} \cup \pi_{result} \cup \pi_{card}$$

*and $\Psi = \langle P_1, P_2 \rangle$ a belief profile. Then,*

$$SE(P_1 *_c P_2) = \{\rho(S) \mid S \in AS(P_{card}[\Psi])\}.$$

5.2.2 *Set-based Revision.* We continue with set-based revision. Unfortunately, the encodings now become more cumbersome, since we cannot make use of weak constraints anymore (which implicitly compared different models and SE models). Instead, we have to use a certain saturation technique (also called *spoiling*), which is quite common to encode problems from the second level of the polynomial hierarchy [Eiter and Gottlob 1995; Eiter and Polleres 2006]. Let us first look at the two main modules for revision $\pi_{witness}$ and $\pi_{incl}$: $\pi_{witness}$ guesses a witness and contains the so-called "spoiling rules" (explained below); this module is also used later in the encodings for merging. $\pi_{incl}$ contains the specific conditions for spoiling in terms the revision operator. Two predicates used in $\pi_{incl}$, $violated$ and $samediff\_all$, will be explained in detail later.

Intuitively, the module $\pi_{witness}$ works as follows. For each interpretation guessed in $\pi_{models}$ we now guess possible witnesses via predicates $win$ and $wout$. Note that this guess (see Rule (18) below) is done via disjunction. This allows us to treat wrong guesses for witnesses via saturation (rather than via constraints as we did in $\pi_{models}$). In fact, such wrong guesses will be indicated via the predicate $spoil$ ($\pi_{incl}$ will provide us with rules which derive $spoil$). Finally, if all such guesses for witnesses are violating some properties, we know that our initial guess (via $in$ and $out$) meets the expected criteria, and the constraint $\leftarrow \sim spoil$ guarantees that only those initial guesses are contained in the answer sets of our encodings. Let us look at this concept in more detail.

DEFINITION 22.

$$\pi_{witness} = \{win(P, A, M)\,;\; wout(P, A, M) \leftarrow prog(P), dom(A), model(M), \tag{18}$$
$$wdiff(P, Q, A, M) \leftarrow win(P, A, M), wout(Q, A, M), \tag{19}$$
$$wdiff(P, Q, A, M) \leftarrow wout(P, A, M), win(Q, A, M), \tag{20}$$
$$wsame(P, Q, A, M) \leftarrow win(P, A, M), win(Q, A, M), \tag{21}$$
$$wsame(P, Q, A, M) \leftarrow wout(P, A, M), wout(Q, A, M), \tag{22}$$
$$notsubseteq(M, I, J) \leftarrow same(P, Q, A, M), wdiff(P, Q, A, M), \tag{23}$$
$$win(P, A, M) \leftarrow spoil, prog(P), dom(A), model(M), \tag{24}$$

---

[8]One might ask why we use the different concepts of $t$- and $c$-models. The reason is that there might be a minimal difference between $(X_1, Y_1)$ and $(X_2, Y_2)$ although there is no minimal difference between $Y_1$ and $Y_2$. But then we still need those interpretations $Y$ in order to compute the corresponding interpretations $X$. On the other hand, there might be a minimal distance between $Y_1$ and $Y_2$ but not between any $(X_1, Y_1)$ and $(X_2, Y_2)$. Still, we then want $(Y_2, Y_2)$ in the result.

$$wout(P, A, M) \leftarrow spoil, prog(P), dom(A), model(M), \qquad (25)$$
$$\leftarrow \sim spoil\}. \qquad (26)$$

As mentioned before, Rule (18) formulates the disjunctive guess for witnesses (referred to as "wguess" in the following) and Rules (19)–(22) provide us with some auxiliary predicates in exactly the same spirit as Rules (6)–(9) from $\pi_{models}$. Rule (23) indicates that the wguess cannot be in a certain subset relation to the initial guess.

Let us illustrate this for the case of revision and suppose the initial guess (via $in$ and $out$) talks about models $M_1$ of $P_1$ and $M_2$ of $P_2$ and the wguess (via $win$ and $wout$) talks about models $N_1$ of $P_1$ and $N_2$ of $P_2$. In case $N_1 \ominus N_2 \nsubseteq M_1 \ominus M_2$, we derive here $notsubseteq(1, 2, c)$.

Before discussing Rules (24)–(26), let us now turn to the specific conditions under which predicate $spoil$ is derived:

DEFINITION 23.

$$\pi_{incl} = \{ spoil \leftarrow win(P, A, h), wout(P, A, t), \qquad (27)$$
$$spoil \leftarrow violated(P, R, M), \qquad (28)$$
$$spoilcond(M) \leftarrow notsubseteq(M, 1, 2), prog\_model(M), \qquad (29)$$
$$spoilcond(c) \leftarrow samediff\_all(c), \qquad (30)$$
$$spoilcond(t) \leftarrow notsubseteq(h, 1, 2), samediff\_all(t), \qquad (31)$$
$$spoilcond(t) \leftarrow samediff\_all(h), samediff\_all(t), \qquad (32)$$
$$spoil \leftarrow spoilcond(c), spoilcond(t), \qquad (33)$$
$$selector(2) \leftarrow \}. \qquad (34)$$

Rule (27) mirrors Rule (5) from $\pi_{models}$ in order to obtain valid SE interpretations. Rule (28) checks whether some wguess does not satisfy some rule (predicate $violated$ is defined in program $\pi_{violation}$ below). Both rules derive predicate $spoil$.

Let us for the moment return to Rules (24) and (25), which derive all possible $win$ and $wout$ predicates in case $spoil$ was derived. Thus, in other words, a wguess which violates some property has to carry to predicate $spoil$, but in case $spoil$ is derived all possible $win$ and $wout$ predicates have to be derived, hence all these wguesses lead to a single answer-set candidate $I$ containing all $win$ and $wout$ predicates (the $in$ and $out$ predicates remain unchanged, however, and thus still characterise an initial guess $G$). In the end, we want that to check whether each wguess violates some property (i.e., carries the $spoil$ predicate). To guarantee that this is the case, Rule (26) finally kills such interpretations, where a wguess would have been valid. Now, due to the minimality of the answer set semantics, if such a wguess existed (for an initial guess $G$), then the set $I$ (containing the same initial guess) cannot become an answer set, and this initial guess has to be withdrawn. In turn, if no wguess without the $spoil$ predicate is left (for an initial guess $G$), then the saturated interpretation $I$ for this initial guess becomes an answer set.

So far, we just have eliminated wguesses which violate the concept of being models and SE models of the given belief profile, respectively. We now have to take the conditions of Definition 8 into account. To this end, we derive $spoil$ for wguesses which are not in the desired relation to the initial guess. Let us first look into the classical models and suppose the initial guess (via $in$ and $out$) makes reference about models $M_1$ of $P_1$ and $M_2$ of $P_2$ and the wguess (via $win$ and $wout$) refers to models $N_1$ of $P_1$ and $N_2$ of $P_2$. In case $N_1 \ominus N_2 \not\subseteq M_1 \ominus M_2$, we already observed that we obtain $notsubseteq(1,2,c)$ and we have one reason to spoil this wguess (cf. Rule (29)). The same holds for $N_1 \ominus N_2 = M_1 \ominus M_2$ (this is done by Rule (30); for the definition of predicate $samediff\_all$, see below). A second reason to spoil is if also the initial guess for the SE models $S_1 = (X_1, Y_1)$ of $P_1$ and $S_2 = (X_2, Y_2)$ of $P_2$ is in a certain relation to the SE model $S'_1 = (X'_1, Y'_1)$ of $P_1$ and $S'_2 = (X'_2, Y'_2)$ of $P_2$ which are given by the wguess. In accordance to Definition 2, we first have to compare the there-parts. Thus, if $Y'_1 \ominus Y'_2 \not\subseteq Y_1 \ominus Y_2$, we already have a reason to spoil (again, taken care of by Rule (29)). If $Y'_1 \ominus Y'_2 = Y_1 \ominus Y_2$, we are not allowed to spoil yet, but have to look into the here-parts. Now, we spoil if $X'_1 \ominus X'_2 \not\subseteq X_1 \ominus X_2$ (Rule (31)) or if $X'_1 \ominus X'_2 = X_1 \ominus X_2$ (Rule (32)). If both reasons for spoil are fulfilled, we finally derive $spoil$ by Rule (33). Finally, rule $selector(2) \leftarrow$ plays the same role as in $\pi_{card}$ and triggers the $\pi_{result}$ module.

It remains to define the predicates $violated$ and $samediff\_all$. One particular problem is that the employed saturation technique allows only for a restricted use of negation. Therefore, we have to re-implement concepts in a different way. In fact, predicates $samediff\_all$ and $violated$ are obtained via additional predicates which loop over all atoms (we now require the concepts defined in module $\pi_{order}$).

Let us have a closer look on the definition of the $violated$ predicate, which derives $violated(P, R, M)$ if the rule $R$ of program $P$ is violated by the current wguess (either in terms of $c$-, $t$-, or $h$-models, as explained earlier). The main idea is to loop over all domain elements. To this end, we first check whether the current guess potentially violates the rule $R$ if we just look at an atom $A$. For instance, if we consider a rule $a \leftarrow b$, and $a$ is in the guess, then this guess cannot violate the rule; the same holds if $b$ is not in the guess. In general, these conditions are formulated by the $unsat$ predicate defined below. Next, we loop over all atoms, and in case we can violate the rule if we take all domain elements $B$ into account ($violupto(P, R, B, M)$), and $A$ is the next element and also allows for violation, we derive $violupto(P, R, A, M)$. rule $R$ is violated by the wguess.

DEFINITION 24.

$$
\begin{aligned}
\pi_{violation} = \{ & unsat(P, R, A, M) \leftarrow win(P, A, M), \sim phead(P, R, A), \\
& \qquad \sim nbody(P, R, A), prog\_rule(P, R), \\
& \qquad prog\_model(M), \\
& unsat(P, R, A, M) \leftarrow wout(P, A, M), \sim pbody(P, R, A), \\
& \qquad \sim nhead(P, R, A), prog\_rule(P, R), \\
& \qquad prog\_model(M), \\
& unsat(P, R, A, h) \leftarrow wout(P, A, h), win(P, A, t), \sim pbody(P, R, A), \\
& \qquad \sim nbody(P, R, A), prog\_rule(P, R), \\
& unsat(P, R, A, h) \leftarrow wout(P, A, h), wout(P, A, t), \sim pbody(P, R, A), \\
& \qquad \sim nhead(P, R, A), prog\_rule(P, R), \\
& violupto(P, R, A, M) \leftarrow inf(A), unsat(P, R, A, M), \\
& violupto(P, R, A, M) \leftarrow succ(B, A), violupto(P, R, B, M), \\
& \qquad unsat(P, R, A, M), \\
& violated(P, R, M) \leftarrow sup(A), violupto(P, R, A, M) \}.
\end{aligned}
$$

The test for equality follows a similar idea. In fact, what we want to do here is, given two sequences of models, $\langle M_\alpha, \ldots, M_n \rangle$ and $\langle N_\alpha, \ldots, N_n \rangle$, to check whether $M_i \ominus M_j = N_i \ominus N_j$ for all $\alpha \le i < j \le n$. The first five rules in the module below do this for any pair of indices $i, j$. By transitivity of $=$, it is then sufficient to simply loop over all indices and check whether the above test holds for all $i, j$ such that $\alpha \le i < n$ and $j = i + 1$.

DEFINITION 25.

$$
\begin{aligned}
\pi_{eq} = \{ & samediff\_atom(I, J, A, M) \leftarrow same(I, J, A, M), wsame(I, J, A, M), \\
& samediff\_atom(I, J, A, M) \leftarrow diff(I, J, A, M), wdiff(I, J, A, M), \\
& samediff\_upto\_atom(I, J, A, M) \leftarrow inf(A), samediff\_atom(I, J, A, M), \\
& samediff\_upto\_atom(I, J, A, M) \leftarrow succ(B, A), samediff\_atom(I, J, A, M), \\
& \qquad\qquad\qquad\qquad\qquad\qquad\qquad samediff\_upto\_atom(I, J, B, M), \\
& samediff(I, J, M) \leftarrow sup(A), samediff\_upto\_atom(I, J, A, M), \\
& samediff\_upto\_prog(I, M) \leftarrow minprog(I), \\
& samediff\_upto\_prog(I, M) \leftarrow I = J + 1, samediff(I, J, M), \\
& \qquad\qquad\qquad\qquad\qquad\qquad samediff\_upto\_prog(J, M), \\
& samediff\_all(M) \leftarrow samediff\_upto\_prog(I, M), maxprog(I) \}.
\end{aligned}
$$

We have the following result.

THEOREM 15. *Let* $\Psi = \langle P_1, P_2 \rangle$ *a belief profile and define*

$P_{incl} = \pi_{domain} \cup \pi_{models} \cup \pi_{order} \cup \pi_{result} \cup \pi_{witness} \cup \pi_{incl} \cup \pi_{violation} \cup \pi_{eq}.$

*Then,*

$$SE(P_1 * P_2) = \{ \rho(S) \mid S \in AS(P_{incl}[\Psi]) \},$$

*where* $\rho$ *is defined as in Theorem 14.*

### 5.3  Encodings for Merging

5.3.1  *Basic Merging.* We continue with the problem of basic merging. We now suppose that belief profiles $\Psi = \langle P_0, \ldots, P_n \rangle$ with an arbitrary number of programs are given. Also recall that $P_0$ plays a special role in basic merging. In particular, the SE models of the result of the merging are taken from the SE models of $P_0$.

In view of the general definition of the previous modules, we do not need to add any further concepts and just define the spoiling conditions:

DEFINITION 26.

$$
\begin{aligned}
\pi_{basic} = \{ & spoil \leftarrow win(P, A, h), wout(P, A, t), \\
& spoil \leftarrow violated(P, R, M), \\
& spoilcond(M) \leftarrow notsubseteq(M, 0, J), prog\_model(M), \\
& spoilcond(c) \leftarrow samediff\_all(c), \\
& spoilcond(t) \leftarrow notsubseteq(h, 0, J), samediff\_all(t), \\
& spoilcond(t) \leftarrow samediff\_all(h), samediff\_all(t), \\
& spoil \leftarrow spoilcond(c), spoilcond(t), \\
& selector(0) \leftarrow, \quad prog(0) \leftarrow \}.
\end{aligned}
$$

Note that we specified $selector(0)$ to select the program from which $\pi_{result}$ takes the models and SE models to define the results, respectively. Also note the fact $prog(0)$ which is necessary in case $P_0$ is the empty program.

We have the following result:

THEOREM 16. *Let* $\Psi = \langle P_0, \dots, P_n \rangle$ *a belief profile and define*

$$P_{basic} = \pi_{domain} \cup \pi_{models} \cup \pi_{order} \cup \pi_{result} \cup \pi_{witness} \cup \pi_{violation} \cup \pi_{eq} \cup \pi_{basic}.$$

*Then,*

$$SE(\Delta(\Psi)) = \{\rho(S) \mid S \in AS(P_{basic}[\Psi])\},$$

*where* $\rho$ *is defined as in Theorem 14.*

5.3.2 *Arbitration Merging.* Our final encoding is the one for arbitration merging. The spoiling module $\pi_{arbitration}$ given next is very much in the style of $\pi_{basic}$ and $\pi_{incl}$. However, we need a somewhat more complicated module to prepare the $resultH$ and $resultT$ predicates, since arbitration merging collects SE models from *all* programs of the belief profile rather than from a single program (which was the case in the approaches we encoded so far). We thus do not use a $selector$ predicate here but instead provide a new result module below. Also recall that belief profiles for arbitration merging are of the form $\langle P_1, \dots, P_n \rangle$. In fact, the only main difference between $\pi_{basic}$ and $\pi_{arbitration}$ is to check that our guess is minimal in a global sense (i.e., the models vary in a minimal way among each other, cf. Definition 12) while in the case of basic merging the minimality has been guaranteed between the designated program $P_0$ and the other programs (cf. Definition 14). This particular difference is reflected by the usage of predicate $notsubseteq(\cdot, I, J)$ compared to $notsubseteq(\cdot, 0, J)$ as used in $\pi_{basic}$.

DEFINITION 27.

$$\begin{aligned}
\pi_{arbitration} = \{ &spoil \leftarrow win(P, A, h), wout(P, A, t), \\
&spoil \leftarrow violated(P, R, A, M), \\
&spoilcond(M) \leftarrow notsubseteq(M, I, J), prog\_model(M), \\
&spoilcond(c) \leftarrow samediff\_all(c), \\
&spoilcond(t) \leftarrow notsubseteq(h, I, J), samediff\_all(t), \\
&spoilcond(t) \leftarrow samediff\_all(h), samediff\_all(t), \\
&spoil \leftarrow spoilcond(c), spoilcond(t), \\
&win(P, A, M) \leftarrow spoil, prog(P), dom(A), model(M), \\
&wout(P, A, M) \leftarrow spoil, prog(P), dom(A), model(M), \\
&\leftarrow \sim spoil \}.
\end{aligned}$$

Here is the new result module:

DEFINITION 28.

$$\begin{aligned}
\pi'_{result} = \{ &tout(I); tout(J) \leftarrow prog(I), prog(J), I \neq J, \\
&tselect(I) \leftarrow \sim tout(I), prog(I), \\
&total \leftarrow \sim nontotal, \\
&nontotal \leftarrow \sim total,
\end{aligned}$$

$$resultT(A) \leftarrow in(M, A, c), tselect(M),$$
$$resultH(A) \leftarrow in(M, A, c), total, tselect(M),$$
$$hout(I); hout(J) \leftarrow prog(I), prog(J), I \neq J, nontotal,$$
$$hselect(I) \leftarrow \sim hout(I), prog(I), nontotal,$$
$$\leftarrow nontotal, in(I, A, t), out(J, A, c), tselect(J), hselect(I), nontotal,$$
$$\leftarrow nontotal, out(I, A, t), in(J, A, c), tselect(J), hselect(I), nontotal,$$
$$resultH(A) \leftarrow in(I, A, h), hselect(I), nontotal\}.$$

Roughly speaking, the first two rules select exactly one program $P_i$ from the belief profile. We then guess whether we build a total or a non-total SE model (as we did in $\pi_{result}$). Then, we copy the model from the guessed program into the there-part of the result, and in case we are constructing a total SE model, also in the here-part. If we construct a non-total SE model, we guess a second program $P_j$ from the belief profile and check whether the there-part of the current SE model of $P_j$ coincides with the classical model of $P_i$ (this is done by the two constraints). If this check is passed, we copy the here-part of the SE model of $P_j$ into the here-part of the resulting SE model.

We arrive at our final result:

THEOREM 17. *Let* $\Psi = \langle P_1, \ldots, P_n \rangle$ *a belief profile and define*

$$P_{arbitration} = \pi_{domain} \cup \pi_{models} \cup \pi_{order} \cup \pi'_{result} \cup$$
$$\pi_{witness} \cup \pi_{violation} \cup \pi_{eq} \cup \pi_{arbitration}.$$

*Then,*

$$SE(\nabla(\Psi)) = \{\rho(S) \mid S \in AS(P_{incl}[\Psi])\},$$

*where* $\rho$ *is defined as in Theorem 14.*

All encodings presented here can be downloaded under the following URL:

http://www.cs.uni-potsdam.de/~torsten/

## 6. DISCUSSION

We have addressed the problem of belief change in logic programming under the answer set semantics. Our overall approach is based on a monotonic characterisation of logic programs, given in terms of the set of SE models of a program. Based on the latter, we first defined and examined operators for logic program expansion and revision. Both subset-based revision and cardinality-based revision were considered. As well as giving properties of these operators, we also considered their complexity. This work is novel, in that it addresses belief change in terms familiar to researchers in belief revision: expansion is characterised in terms of intersections of models, and revision is characterised in terms of minimal distance between models.

We also addressed the problem of merging logic programs under the answer set semantics. Again, the approaches are based on a monotonic characterisation of logic programs, given in terms of the set of SE models of a sequence of programs. We defined and examined two operators for logic program merging, the first following intuitions from arbitration [Liberatore and Schaerf 1998], the second being closer to IC merging [Konieczny and Pino Pérez 2002]. Notably, since these merging operators are defined via a semantic characterisation, the results of merging are independent of the particular syntactic expression of a logic program. As well as giving properties of these operators, we also considered complexity questions. Last, we provided encodings for both program revision and program merging.

We note that previous work on logic program belief change was formulated at the level of the individual program, and not in terms of an abstract characterisation (via strong equivalence or sets of SE interpretations). The net result is that such previous work is generally difficult to work with: properties are difficult to come by, and often desirable properties (such as the tautology property) are lacking. The main point of departure for the current approach then is to lift the problem of logic program revision or merging from the program (or syntactic) level to an abstract (or semantic) level.

## A.   APPENDIX

### A.1   Proof of Theorem 1

Most of the parts follow immediately from the fact that $SE(P + Q) = SE(P) \cap SE(Q)$.

(1)   We need to show that Definition 6 results in a well-defined set of SE models.

For $SE(P) \cap SE(Q) = \emptyset$ we have that $\emptyset$ is trivially well-defined (and $R$ can be given by $\bot \leftarrow$).

Otherwise, for $SE(P) \cap SE(Q) \neq \emptyset$, we have the following: If $(X, Y) \in SE(P) \cap SE(Q)$, then $(X, Y) \in SE(P)$ and $(X, Y) \in SE(Q)$; whence $(Y, Y) \in SE(P)$ and $(Y, Y) \in SE(Q)$ since $SE(P)$ and $SE(Q)$ are well-defined by virtue of $P$ and $Q$ being logic programs. Hence, $(Y, Y) \in SE(P) \cap SE(Q)$. Since this holds for arbitrary $(X, Y) \in SE(P) \cap SE(Q)$, we have that $SE(P) \cap SE(Q)$ is well-defined.

(2)   Immediate from the definition of $+$.

(3)   If $P \models_s Q$, then $SE(P) \subseteq SE(Q)$. Hence, $SE(P) \cap SE(Q) = SE(P)$, or $P + Q \equiv_s P$.

(4)   Similar to the previous part.

(5)   This was established in the first part.

(6)   To show completeness, we need to show that for any $(X, Y) \in SE(P + Q)$ and $(Y \cup Y', Y \cup Y') \in SE(P + Q)$ that $(X, Y \cup Y') \in SE(P + Q)$.

If $(X, Y) \in SE(P + Q)$ and $(Y \cup Y', Y \cup Y') \in SE(P + Q)$, then $(X, Y) \in SE(P) \cap SE(Q)$ and $(Y \cup Y', Y \cup Y') \in SE(P) \cap SE(Q)$. Hence, $(X, Y) \in SE(P)$ and $(Y \cup Y', Y \cup Y') \in SE(P)$, and so, since $SE(P)$ is complete by assumption, we have $(X, Y \cup Y') \in SE(P)$.

The same argument gives that $(X, Y \cup Y') \in SE(Q)$, whence $(X, Y \cup Y') \in SE(P) \cap SE(Q)$ and $(X, Y \cup Y') \in SE(P + Q)$.

(7)   If $Q \equiv_s \emptyset$, then $SE(Q) = \{(X, Y) \mid X \subseteq Y \subseteq \mathcal{A}\}$ from which the result follows immediately.   □

### A.2   Proof of Theorem 2

($RA1$). This postulate follows immediately from Definition 8. Note that $(X, Y) \in SE(P * Q)$ only if $Y \in \sigma(Mod(Q), Mod(P))$, and therefore $(Y, Y) \in \sigma(SE(Q), SE(P))$. So, $SE(P * Q)$ is well-defined.

($RA2$). If $P + Q$ is satisfiable, then we have that both $\sigma(Mod(Q), Mod(P)) \neq \emptyset$ and $\sigma(SE(Q), SE(P)) \neq \emptyset$. Further, for $Y \in Mod(Q)$ (or $(X, Y) \in SE(Q)$), there is some $Y' \in Mod(P)$ (resp., $(X', Y') \in SE(P)$) such that $Y \ominus Y' = \emptyset$ (resp., $(X, Y) \ominus (X', Y') = \emptyset$), from which our result follows.

$(RA3)$. From Definition 8 we have that, if $P$ is unsatisfiable, then $Q$ is satisfiable iff $P * Q$ is satisfiable.

Otherwise, if $P$ is satisfiable and $Q$ is satisfiable, then there is some $(Y, Y) \in \sigma(Mod(Q), Mod(P))$ (since $SE(Q)$ is well-defined and given Definition 7). Hence, $SE(P * Q) \neq \emptyset$.

$(RA4)$. Immediate from Definition 8.

$(RA5)$. If $SE(P) = \emptyset$, then the result follows immediately from the first part of Definition 8.

Otherwise, we show that, if $(X, Y)$ is an SE model of $(P * Q) + R$, then $(X, Y)$ is an SE model of $P * (Q + R)$.

Let $(X, Y) \in SE((P * Q) + R)$. Then, $(X, Y) \in SE(P * Q)$ and $(X, Y) \in SE(R)$. Since $(X, Y) \in SE(P * Q)$, by $(RA1)$ we have that $(X, Y) \in SE(Q)$, and so $(X, Y) \in SE(Q) \cap SE(R)$, or $(X, Y) \in SE(Q + R)$.

There are two cases to consider:

$X = Y$: Since then $(X, Y) = (Y, Y)$, and $(Y, Y) \in SE(P * Q)$, we have that $Y \in \sigma(Mod(Q), Mod(P))$. Hence, from Definition 7, $Y \in Mod(Q)$ and there is some $Y' \in Mod(P)$ such that there is no $Y_1 \in Mod(Q)$ and no $Y_2 \in Mod(P)$ such that $Y_1 \ominus Y_2 \subset Y \ominus Y'$.

We established at the outset that $(X, Y) \in SE(Q + R)$. Hence, $Y \in Mod(Q + R)$. This gives us that $Y \in Mod(Q + R)$ and there is some $Y' \in Mod(P)$ such that no $Y_1, Y_2$ exist with $Y_1 \in Mod(Q)$, $Y_2 \in Mod(P)$, and $Y_1 \ominus Y_2 \subset Y \ominus Y'$.

Clearly, in the above, if there is no $Y_1 \in Mod(Q)$ such that the above condition holds, then there is no $Y_1 \in Mod(Q + R)$ such that the above condition holds.

Thus, we have $Y \in Mod(Q + R)$ and there is some $Y' \in Mod(P)$ for which no $Y_1 \in Mod(Q + R)$ and no $Y_2 \in Mod(P)$ exists such that $Y_1 \ominus Y_2 \subset Y \ominus Y'$.

Thus, from Definition 7, we get $Y \in \sigma(Mod(Q + R), Mod(P))$, hence $(Y, Y) \in SE(P * (Q + R))$.

$X \subset Y$: We have $Y \in \sigma(Mod(Q), Mod(P))$ by virtue of $(X, Y) \in SE(P * Q)$. In the previous part we established that $Y \in \sigma(Mod(Q + R), Mod(P))$.

As well, $(X, Y) \in \sigma(SE(Q), SE(P))$ since $(X, Y) \in SE(P * Q)$. Thus, from Definition 7, we have that there is some $(X', Y') \in SE(P)$ such that no $U, V, U', V'$ exist such that $(U, V) \in SE(Q)$, $(U', V') \in SE(P)$, and $(U, V) \ominus (U', V') \subset (X, Y) \ominus (X', Y')$. Therefore, there is no $(U, V) \in SE(Q + R)$ and no $(U', V') \in SE(P)$ such that $(U, V) \ominus (U', V') \subset (X, Y) \ominus (X', Y')$.

We previously showed that $(X, Y) \in SE(Q + R)$. Consequently, from Definition 8, we obtain that $(X, Y) \in \sigma(SE(Q + R), SE(P))$. Hence, $(X, Y) \in SE(P * (Q + R))$.

Thus, in either case, we get $(X, Y) \in SE(P * (Q + R))$, which was to be shown. $\square$

### A.3   Proof of Theorem 3

For initialisation, idempotency, and tautology, in the left-hand side of the given equivalence, revision corresponds with expansion via $(RA2)$, from which the result is immediate.

For absorption, we have $Q = R$, and so $((P * Q) * R) = ((P * Q) * Q)$. Since $SE(P * Q) \subseteq SE(Q)$, then from Theorem 1, Part 3, we have that $(P * Q) + Q \equiv_s P * Q$. As well, $((P * Q) * Q) = ((P * Q) + Q)$, from which our result follows. $\square$

### A.4 Proof of Theorem 4

We need to show that $SE(P * Q) \subseteq SE(P *_w Q)$. If $SE(P) = \emptyset$, then $SE(P * Q) = SE(Q) = SE(P *_w Q)$.

Otherwise, there are two cases to consider:

(1) $(X, Y) \in SE(P * Q)$ where $X \subset Y$. Then, $(X, Y) \in \sigma(SE(P), SE(Q))$ by Definition 8, and $(X, Y) \in SE(P *_w Q)$ by Definition 9.

(2) $(Y, Y) \in SE(P * Q)$. From Definition 8, we have that $Y \in \sigma(Mod(Q), Mod(P))$. $Y \in \sigma(Mod(Q), Mod(P))$ implies that $(Y, Y) \in \sigma(SE(Q), SE(P))$. Hence, according to Definition 9, $(Y, Y) \in SE(P *_w Q)$.

Therefore, $(X, Y) \in SE(P * Q)$ implies that $(X, Y) \in SE(P *_w Q)$, whence $SE(P * Q) \subseteq SE(P *_w Q)$. □

### A.5 Proof of Theorem 5

Before giving the proof, we first present a lemma that is key for postulates $(RA5)$ and $(RA6)$.

LEMMA 1. *Let $E_1$, $E_2$, and $E_3$ be SE interpretations. If $\sigma_{||}(E_1, E_2) \cap E_3 \neq \emptyset$, then $\sigma_{||}(E_1, E_2) \cap E_3 = \sigma_{||}(E_1 \cap E_3, E_2)$.*

PROOF. Assume that $\sigma_{||}(E_1, E_2) \cap E_3 \neq \emptyset$.

For showing $\subseteq$ in the equality, let $(X, Y) \in \sigma_{||}(E_1, E_2) \cap E_3$ and, toward a contradiction, assume that $(X, Y) \notin \sigma_{||}(E_1 \cap E_3, E_2)$.

Since $(X, Y) \in \sigma_{||}(E_1, E_2)$, so $(X, Y) \in E_1$; as well, $(X, Y) \in E_3$, so $(X, Y) \in E_1 \cap E_3$. Since $(X, Y) \notin \sigma_{||}(E_1 \cap E_3, E_2)$ we have that there is some $(X', Y') \in E_1 \cap E_3$ and some $(U', V') \in E_2$ such that for every $(U, V) \in E_2$, $|(X', Y') \ominus (U', V')| < |(X, Y) \ominus (U, V)|$. But this contradicts the assumption that $(X, Y) \in \sigma_{||}(E_1, E_2)$. Hence, the assumption that $(X, Y) \notin \sigma_{||}(E_1 \cap E_3, E_2)$ cannot hold, i.e., $(X, Y) \in \sigma_{||}(E_1 \cap E_3, E_2)$, establishing that $\sigma_{||}(E_1, E_2) \cap E_3 \subseteq \sigma_{||}(E_1 \cap E_3, E_2)$.

To show $\supseteq$ in the equality, let $(X, Y) \in \sigma_{||}(E_1 \cap E_3, E_2)$ and, toward a contradiction, assume that $(X, Y) \notin \sigma_{||}(E_1, E_2) \cap E_3$.

Since $(X, Y) \in \sigma_{||}(E_1 \cap E_3, E_2)$, we get that $(X, Y) \in E_3$. Hence, $(X, Y) \notin \sigma_{||}(E_1, E_2)$ (via the assumption that $(X, Y) \notin \sigma_{||}(E_1, E_2) \cap E_3$).

We also have by assumption that $\sigma_{||}(E_1, E_2) \cap E_3 \neq \emptyset$, and so let $(X', Y') \in \sigma_{||}(E_1, E_2) \cap E_3$. Then, from the first part above, we have that $(X', Y') \in \sigma_{||}(E_1 \cap E_3, E_2)$. Thus, we have both that $(X, Y) \in \sigma_{||}(E_1 \cap E_3, E_2)$ and $(X', Y') \in \sigma_{||}(E_1 \cap E_3, E_2)$. Consequently, we obtain that

$$min(\{|(X, Y) \ominus (U, V)| \mid (U, V) \in E_2\}) = min(\{|(X', Y') \ominus (U, V)||(U, V) \in E_2\}).$$

Therefore, since $(X', Y') \in \sigma_{||}(E_1, E_2)$, so also $(X, Y) \in \sigma_{||}(E_1, E_2)$. But this together with $(X, Y) \in E_3$ contradicts our assumption that $(X, Y) \notin \sigma_{||}(E_1, E_2) \cap E_3$; i.e., we have $(X, Y) \in \sigma_{||}(E_1, E_2) \cap E_3$, establishing that $\sigma_{||}(E_1, E_2) \cap E_3 \supseteq \sigma_{||}(E_1 \cap E_3, E_2)$. □

We now move on to the proof of Theorem 5:

$(RA1)$. This follows immediately from Definition 11. Note that $(X, Y) \in SE(P *_c Q)$ only if $Y \in \sigma_{||}(Mod(Q), Mod(P))$, and therefore $(Y, Y) \in \sigma_{||}(SE(Q), SE(P))$. So, $SE(P *_c Q)$ is well-defined.

$(RA2)$. If $P + Q$ is satisfiable, then we have that both $\sigma_{||}(Mod(Q), Mod(P)) \neq \emptyset$ and $\sigma_{||}(SE(Q), SE(P)) \neq \emptyset$. Further, for $Y \in Mod(Q)$ (or $(X,Y) \in SE(Q)$) we have that there is some $Y' \in Mod(P)$ (resp., $(X',Y') \in SE(P)$) such that $Y \ominus Y' = \emptyset$ $((X,Y) \ominus (X',Y') = \emptyset)$, from which our result follows.

$(RA3)$. From Definition 11 we have that, if $P$ is unsatisfiable, then $Q$ is satisfiable iff $P * Q$ is satisfiable. Otherwise, if $P$ is satisfiable and $Q$ is satisfiable, then there is some $(Y,Y) \in \sigma_{||}(Mod(Q), Mod(P))$ (since $SE(Q)$ is well-defined and given Definition 10). Hence, $SE(P * Q) \neq \emptyset$.

$(RA4)$. This is immediate from Definition 11.

$(RA5)$, $(RA6)$. For $P *_c Q$, if $SE(P) = \emptyset$, we have that $(P *_c Q) + R = Q + R = (P *_c (Q + R))$.

So, assume that $SE(P) \neq \emptyset$. We show that $SE(P *_c Q) + SE(R) = SE(P *_c (Q+R))$, thus establishing both postulates.

For $\subseteq$, assume that $(X,Y) \in SE(P *_c Q) + R$. Thus, $(X,Y) \in SE(P *_c Q)$ and $(X,Y) \in R$.

For $X \subseteq Y$, we have that $Y \in \sigma_{||}(Mod(Q), Mod(P))$ and as well $Y \in Mod(R)$. We get that $Y \in \sigma_{||}(Mod(Q+R), Mod(P))$ by the analogous proof in propositional logic for cardinality-based revision.

For $X \subset Y$, we have that $(X,Y) \in \sigma_{||}(SE(Q), SE(P))$ and as well $(X,Y) \in SE(R)$. By Lemma 1 we get that $(X,Y) \in \sigma_{||}(SE(Q+R), SE(P))$.

This establishes one direction of the set equality. For $\supseteq$, the argument is essentially the same, though in the reverse direction, and again appealing to Lemma 1.

## A.6    Proof of Theorem 6

The proof is the same as for Theorem 3.

## A.7    Proof of Theorem 7

Since we deal with a globally fixed language, we first need a few lemmata.

LEMMA 2. *Let $P, Q$ be programs, $Y$ an interpretation, and $x \in Y \setminus var(P \cup Q)$. Then, $Y \in \sigma(Mod(Q), Mod(P))$ implies $Y \setminus \{x\} \in \sigma(Mod(Q), Mod(P))$.*

PROOF. Since $Y \in \sigma(Mod(Q), Mod(P))$, so $Y \in Mod(Q)$ and there exists some $Z \in Mod(P)$ such that for each $Y' \in Mod(Q)$ and $Z' \in Mod(P)$, $Y' \ominus Z' \not\subset Y \ominus Z$. We show that $x \in Z$ holds. Suppose this is not the case: Then, we have $x \in Y \ominus Z$, since $x \in Y$. Now, since $x \notin var(P)$, also $Z \cup \{x\} \in Mod(P)$. But then $x \notin Y \ominus (Z \cup \{x\})$ which yields $Y \ominus (Z \cup \{x\}) \subset Y \ominus Z$, a contradiction to our assumption. Hence, we can suppose $x \in Z$. Now, since $Y \in Mod(Q)$, obviously $Y \setminus \{x\} \in Mod(Q)$ as well. We obtain $Y \ominus Z = (Y \setminus \{x\}) \ominus (Z \setminus \{x\})$, thus $Y \setminus \{x\} \in \sigma(Mod(Q), Mod(P))$ holds.    □

LEMMA 3. *Let $P, Q$ be programs, $(X,Y)$ an SE interpretation, and $x \in Y \setminus var(P \cup Q)$. Then, $(X,Y) \in \sigma(SE(Q), SE(P))$ implies $(X \setminus \{x\}, Y \setminus \{x\}) \in \sigma(SE(Q), SE(P))$.*

PROOF. Since $(X,Y) \in \sigma(SE(Q), SE(P))$, $(X,Y) \in SE(Q)$ and there exists a $(U,Z) \in SE(P)$ such that for each $(X',Y') \in SE(Q)$ and each $(U',Z') \in SE(P)$, $(X',Y') \ominus (U',Z') \not\subset (X,Y) \ominus (U,Z)$. We show that the following relations hold: (i) $x \in Z$; and (ii) $x \in U$ iff $x \in X$. Towards a contradiction, first suppose $x \notin Z$. Then, we have $x \in Y \ominus Z$, since $x \in Y$. Now, since $x \notin var(P)$, also $(U, Z \cup \{x\}) \in SE(P)$ and $(U \cup \{x\}, Z \cup \{x\}) \in SE(P)$. We have $x \notin Y \ominus (Z \cup \{x\})$ which yields $Y \ominus (Z \cup \{x\}) \subset Y \ominus Z$. Thus, $(X, Y) \ominus (U, Z \cup \{x\}) \subset (X, Y) \ominus (U, Z)$, which is a contradiction to the assumption. Hence, $x \in Z$ holds. If (ii) does not hold, we get $x \in X \oplus U$. Now, in case $x \in X$ and $x \notin U$, we have $(X, Y) \ominus (U \cup \{x\}, Z) \subset (X, Y) \ominus (U, Z)$. In case $x \in U$ and $x \notin X$, we have $(X, Y) \ominus (U \setminus \{x\}, Z) \subset (X, Y) \ominus (U, Z)$. Again, both cases yield a contradiction. Clearly, $(X, Y) \in SE(Q)$ implies $(X \setminus \{x\}, Y \setminus \{x\}) \in SE(Q)$ and we obtain $(X, Y) \ominus (U, Z) = (X \setminus \{x\}, Y \setminus \{x\}) \ominus (U \setminus \{x\}, Z \setminus \{x\})$. $(X \setminus \{x\}, Y \setminus \{x\}) \in \sigma(SE(Q), SE(P))$ thus follows. $\square$

LEMMA 4. *For any programs $P$, $Q$, and $R$, $P * Q \not\models_s R$ iff there exist $X \subseteq Y \subseteq var(P \cup Q \cup R)$ such that $(X, Y) \in SE(P * Q)$ and $(X, Y) \notin SE(R)$.*

PROOF. The if-direction is by definition.

As for the only-if direction, assume $P * Q \not\models_s R$. Then, there exists a pair $(X, Y)$ such that $(X, Y) \in SE(P * Q)$ and $(X, Y) \notin SE(R)$. Let $V = var(P \cup Q \cup R)$. We first show that $(X \cap V, Y \cap Y) \in SE(P * Q)$. By definition, $(X, Y) \in SE(Q)$. If $SE(P) = \emptyset$, $SE(P * Q) = SE(Q)$, and since $(X, Y) \in SE(Q)$ obviously implies $(X \cap V, Y \cap Y) \in SE(Q)$, $(X \cap V, Y \cap Y) \in SE(P * Q)$ thus follows in this case. So suppose $SE(P) \neq \emptyset$. Then, $Y \in \sigma(Mod(Q), Mod(P))$. By iteratively applying Lemma 2, we obtain that also $Y \cap V \in \sigma(Mod(Q), Mod(P))$. Analogously using Lemma 3, $(X, Y) \in \sigma(SE(Q), SE(P))$ yields $(X \cap V, Y \cap V) \in \sigma(SE(Q), SE(P))$. By Definition 8, we get $(X \cap V, Y \cap V) \in SE(P * Q)$. Finally, it is clear that $(X, Y) \notin SE(R)$, implies that $(X \cap V, Y \cap V) \notin SE(R)$. $\square$

We now proceed with the proof of Theorem 7.

We first show membership in $\Sigma_2^P$ for the complementary problem. From Lemma 4, the complementary problem holds iff there exist $X, Y \subseteq var(P \cup Q \cup R)$ such that $(X, Y) \in SE(P * Q)$ and $(X, Y) \notin SE(R)$. In what follows, let $V = var(P \cup Q \cup R)$. We first state the following observation: Recall that $Y \in \sigma(Mod(Q), Mod(P))$ iff $Y \in Mod(Q)$ and there exists a $W \in Mod(P)$ such that $W \subseteq V$ and for each $Y' \in Mod(Q)$ and $W' \in Mod(P)$, $Y' \oplus W' \not\subset Y \ominus W$. Now, if $Y \subseteq V$, then there is also a $W \subseteq V$ satisfying above test (this is seen by the arguments used in the proof of Lemma 2). A similar observation holds for $(X, Y) \in \sigma(SE(Q), SE(P))$.

Thus an algorithm to decide $P * Q \not\models_s R$ is as follows. We guess interpretations $X, Y, W, U, Z \subseteq V$ and start with checking $(X, Y) \in SE(Q)$ and $(X, Y) \notin SE(R)$. Then, we check whether $SE(P) = \emptyset$ which can be done via a single call to an NP-oracle. If the answer is yes, we already have found an SE interpretation $(X, Y)$ such that $(X, Y) \in SE(P * Q)$ and $(X, Y) \notin SE(R)$ and thus the complementary problem holds. If the answer is no, we next check whether $(U, Z) \in SE(P)$ and $W \in Mod(P)$. Then, (i) given $Y$ and $W$, we check whether for each $Y' \subseteq V$ and each $W' \subseteq V$ such that $Y' \in Mod(Q)$ and $W' \in Mod(P)$, $Y' \ominus W' \not\subset Y \ominus W$ holds. It is easy to see that then the same relation holds for arbitrary models $Y'$ and $W'$. From that we can conclude that $Y \in \sigma(Mod(Q), Mod(P))$. Next, (ii) given $(X, Y)$ and $(U, Z)$, we check whether for each $X' \subseteq Y' \subseteq V$ and each $U' \subseteq Z' \subseteq V$ such that $(X', Y') \in SE(Q)$ and $(U', W') \in SE(P)$, $(X', Y') \ominus (U', W') \not\subset (X, Y) \ominus (U, W)$. Again, it is easy to see that in this case $(X, Y) \in \sigma(SE(Q), SE(P))$ follows. But then we obtain $(X, Y) \in SE(P * Q)$ by Definition 8 which together with $(X, Y) \notin SE(R)$ solves the complementary problem in view of Lemma 4.

We recall that model checking as well as SE model checking are in P. So most of the checks used above are in P (expect the already mentioned call to an NP-oracle) and it remains to settle the complexity of the checks (i) and (ii). As well, they can be done by an NP-oracle. This can be seen by considering the respective complementary problems, where one guesses the sets $Y', W'$ (resp., $X', Y', U', Z'$) and then performs model checking or SE model checking together with some other simple tests which are all in P. Thus, the overall algorithm runs in nondeterministic polynomial time with access to an NP-oracle. This shows the $\Sigma_2^P$-membership as desired.

As for the hardness-part, we use a reduction from the problem of checking whether a given quantified Boolean formula of form $\Phi = \forall Y \exists X \varphi$, where $\varphi$ is a propositional formula in conjunctive normal form over atoms $X \cup Y$, evaluates to true, which is $\Pi_2^P$-complete. For $\Phi$ as described, let, for each $z \in X \cup Y$, $z'$ be a new atom. Additionally, for each clause $c = z_1 \vee \cdots \vee z_k \vee \neg z_{k+1} \vee \cdots \vee \neg z_m$ in $\varphi$, let $\hat{c}$ be the sequence $z'_1, \ldots, z'_k, z_{k+1}, \ldots, z_m$. Finally, let $w$ be a further new atom and $V = X \cup Y \cup \{z' \mid z \in X \cup Y\} \cup \{w\}$. We define the following programs: $P_\Phi = \{v \leftarrow \mid v \in V\}$, $R_\Phi = \{w \leftarrow\}$, and

$$Q_\Phi = \{y \leftarrow \sim y';\ y' \leftarrow \sim y;\ \bot \leftarrow y, y' \mid y \in Y\} \cup$$
$$\{x \leftarrow \sim x', w;\ x' \leftarrow \sim x, w;\ w \leftarrow x;\ w \leftarrow x';$$
$$\bot \leftarrow x, x' \mid x \in X\} \cup$$
$$\{\bot \leftarrow \hat{c}, w \mid c \text{ a clause in } \varphi\}.$$

The SE models over $V$ of these programs are as follows (for a set $Z$ of atoms, $Z'$ stands for $\{z' \mid z \in Z\}$):

$$SE(P_\Phi) = \{(V, V)\};$$
$$SE(Q_\Phi) = \{(S, S) \mid S = I \cup (Y \setminus I)', I \subseteq Y\} \cup$$
$$\{(S, T), (T, T) \mid S = I \cup (Y \setminus I)',$$
$$T = \{w\} \cup S \cup J \cup (X \setminus J)',$$
$$I \subseteq Y, J \subseteq X, I \cup J \models \varphi\};$$
$$SE(R_\Phi) = \{(W_1, W_2) \mid \{w\} \subseteq W_1 \subseteq W_2 \subseteq V\}.$$

We show that $\Phi$ is true iff $P_\Phi * Q_\Phi \models_s R_\Phi$ holds.

*Only-if direction*: Suppose $P_\Phi * Q_\Phi \models_s R_\Phi$ does not hold. By Lemma 4, there exist $S \subseteq T \subseteq var(P_\Phi \cup Q_\Phi \cup R_\Phi) = V$ such that $(S, T) \in SE(P_\Phi * Q_\Phi)$ and $(S, T) \notin SE(R_\Phi)$. Inspecting the SE models of $R_\Phi$, we obtain that $w \notin S$. From $(S, T) \in SE(P_\Phi * Q_\Phi)$, $(S, T) \in SE(Q_\Phi)$, and thus $S$ has to be of the form $I \cup (Y \setminus I)'$ for some $I \subseteq Y$. Recall that $(V, V)$ is the only SE model of $P_\Phi$ over $V$. Hence, $S = T$ holds, since otherwise $(T, T) \ominus (V, V) \subset (S, T) \ominus (V, V)$, which is in contradiction to $(S, T) \in SE(P_\Phi * Q_\Phi)$. Now we observe that for each $U$ with $S = T \subset U \subseteq V$, $(U, U) \notin SE(Q_\Phi)$ has to hold, (otherwise $(U, U) \ominus (V, V) \subset (S, S) \ominus (V, V)$). Inspecting the SE models of $SE(Q_\Phi)$, this only holds if, for each $J \subseteq X$, $I \cup J \not\models \varphi$. But then $\Phi$ is false.

*If direction*: Suppose $\Phi$ is false. Then, there exists an $I \subseteq Y$ such that for all $J \subseteq X$, $I \cup J \not\models \varphi$. We know that $(S, S) = (I \cup (Y \setminus I)', I \cup (Y \setminus I)') \in SE(Q_\Phi)$ and $(V, V) \in SE(P_\Phi)$. Next, to obtain $(S, S) \in SE(P_\Phi * Q_\Phi)$, we show $S \in \sigma(Mod(Q_\Phi), Mod(P_\Phi))$. Suppose this is not the case. Since $S \subset V$ and $V$ is the minimal model of $P_\Phi$, there has to exist an $U$ with $S \subset U \subseteq V$ such that $U \in Mod(Q_\Phi)$. Recall that $S = I \cup (Y \setminus I)'$ and, by assumption, for all $J \subseteq X$, $I \cup J \not\models \varphi$. By inspecting the SE models of $Q_\Phi$, it is clear that no such $U \in Mod(Q_\Phi)$ exists. By essentially the same arguments, $(S, S) \in \sigma(SE(Q_\Phi), SE(P_\Phi))$ can be shown. Therefore, $(S, S) \in SE(P_\Phi * Q_\Phi)$ and since $w \notin S$, $P_\Phi * Q_\Phi \models_s R_\Phi$ does not hold.

This shows $\Pi_2^P$-hardness for normal programs $Q$. The result for positive programs $Q$ is obtained by replacing in $Q_\Phi$ rules $y \leftarrow \sim y'$, $y' \leftarrow \sim y$ by $y$; $y' \leftarrow$, and likewise rules $x \leftarrow \sim x', w$ and $x' \leftarrow \sim x, w$ by $x$; $x' \leftarrow w$. Due to the presence of the constraints $\bot \leftarrow y, y'$ and $\bot \leftarrow x, x'$, this modification does not change the SE models of these programs.

## A.8   Proof of Theorem 9

The definitions for arbitration and basic merging (Definitions 13 and 15) are essentially composed of two parts (as are the definitions for revision): there is a phrase to deal with classical propositional models (or SE models of form $(Y, Y)$) and then general SE models. For brevity, and because the case for propositional models follows immediately from the case of general SE models, we consider general SE models in the proofs here.

$(LS1') - (LS7')$. These all follow trivially or straightforwardly from the definition of $P_1 \diamond P_2$.

$(LS8')$. Assume that $P_1$ and $P_2$ are satisfiable. It follows that $SE(\langle P_1, P_2 \rangle) \neq \emptyset$ and so $Min_a(SE(\langle P_1, P_2 \rangle)) \neq \emptyset$. Let $\langle S_1, S_2 \rangle \in Min_a(SE(\langle P_1, P_2 \rangle))$, and so $S_1, S_2 \in SE(P_1 \diamond P_2)$. Since $S_1 \in SE(P_1)$ we get that $S_1 \in SE(P_1) \cap SE(P_1 \diamond P_2)$ and so $S_1 \in SE(P_1 \sqcap (P_1 \diamond P_2))$. Thus, $P_1 \sqcap (P_1 \diamond P_2)$ is satisfiable.

## A.9   Proof of Theorem 10

Let $\Psi$ be a belief profile, $P_0$ a program representing global constraints, and $\Delta$ as given in Definition 15. Then, $\Delta$ satisfies the following versions of the postulates of Definition 5:

$(IC0') - (IC3')$, $(IC9')$. These follow trivially or straightforwardly from the definition of $\Delta(\langle P_0, \Psi \rangle)$.

$(IC4')$. Assume that $P_1 \models_s P_0$ and $P_2 \models_s P_0$. If $SE(P_1) \cap SE(P_2) \neq \emptyset$ then by $(IC2')$ we have that $\Delta(\langle P_0, P_1, P_2 \rangle) = P_0 \sqcap P_1 \sqcap P_2$ from which our result follows immediately.

Consequently, assume that $SE(P_1) \cap SE(P_2) = \emptyset$. As well, assume the antecedent condition of the postulate that $\Delta(\langle P_0, P_1, P_2 \rangle) \sqcap P_1$ is satisfiable. Let $\Psi = \langle P_0, P_1, P_2 \rangle$. Thus, we have for some $(X, Y)$ that $(X, Y) \in SE(\Delta(\Psi) \sqcap P_1)$, and so $(X, Y) \in SE(P_0) \cap SE(P_1)$, where $(X, Y) \in Min_b(SE(\Psi))_0$.

$(X, Y) \in Min_b(SE(\Psi))_0$ implies that there is some $(X', Y') \in SE(P_2)$ such that $\overline{S} = \langle (X, Y), (X, Y), (X', Y') \rangle \in Min_b(SE(\Psi))$.

We claim that $\overline{S}' = \langle (X', Y'), (X, Y), (X', Y') \rangle \in Min_b(SE(\Psi))$. This is sufficient to prove our result, since $\overline{S}' \in Min_b(SE(\Psi))$ yields that $(X', Y') \in \Delta(\Psi)$ and $(X', Y') \in SE(P_2)$, that is to say, $\Delta(\Psi) \sqcap P_2$ is satisfiable.

Proof of claim: Since $\overline{S} \in Min_b(SE(\Psi))$, this means that for every $\overline{T} \in SE(\Psi)$ we have that $\overline{T} \leq_b \overline{S}$ implies that $\overline{S} \leq_b \overline{T}$.

Consider $\overline{T} = \langle (U_0, V_0), (U_1, V_1), (U_2, V_2) \rangle$. If $\overline{T} \leq_b \overline{S}$ then we have that $(U_0, V_0) \ominus (U_1, V_1) \subseteq (X, Y) \ominus (X, Y) = (\emptyset, \emptyset)$. That is, $U_0 = U_1$ and $V_0 = V_1$, and so $\overline{T} =$ $\langle(U_0, V_0), (U_0, V_0), (U_2, V_2)\rangle$. As well, from $\overline{T} \leq_b \overline{S}$, we get that $(U_0, V_0) \ominus (U_2, V_2) \subseteq (X, Y) \ominus (X', Y')$. Since $\overline{T} \leq_b \overline{S}$ implies $\overline{S} \leq_b \overline{T}$, this means that $(X, Y) \ominus (X', Y') \subseteq (U_0, V_0) \ominus (U_2, V_2)$.

We will use this later, and so summarise the result here:

($\alpha$) $(X, Y)$ and $(X', Y')$ are such that for every $(U_0, V_0) \in SE(P_1)$ and $(U_2, V_2) \in SE(P_2)$ if $(U_0, V_0) \ominus (U_2, V_2) \subseteq (X, Y) \ominus (X', Y')$ then $(X, Y) \ominus (X', Y') \subseteq (U_0, V_0) \ominus (U_2, V_2)$.

We must show for $\overline{S}' = \langle(X', Y'), (X, Y), (X', Y')\rangle$, that $\overline{T} \leq_b \overline{S}'$ implies $\overline{S}' \leq_b \overline{T}$. Let $\overline{T} = \langle(U_0', V_0'), (U_1', V_1'), (U_2', V_2')\rangle$ and assume that $\overline{T} \leq_b \overline{S}'$. Then, by definition of $\leq_b$, we have that $(U_0', V_0') \ominus (U_1', V_1') \subseteq (X', Y') \ominus (X, Y)$. As well, we have that $(U_0', V_0') \ominus (U_2', V_2') \subseteq (X', Y') \ominus (X', Y') = (\emptyset, \emptyset)$. Hence, we must have that $U_0' = U_2'$ and $V_0' = V_2'$. Thus, we can write $\overline{T} = \langle(U_0', V_0'), (U_1', V_1'), (U_0', V_0')\rangle$.

Now, we will have $\overline{S}' \leq_b \overline{T}$ just if $(X', Y') \ominus (X, Y) \subseteq (U_0', V_0') \ominus (U_1', V_1')$ and $(X', Y') \ominus (X', Y') \subseteq (U_0', V_0') \ominus (U_2', V_2')$. The second condition is vacuously true. As for the first condition, we have that $(U_0', V_0') \in SE(P_2)$ and $(U_1', V_1') \in SE(P_1)$. Thus, via ($\alpha$), we obtain that $(X', Y') \ominus (X, Y) \subseteq (U_1', V_1') \ominus (U_0', V_0')$. We conclude that $\overline{S}' \leq_b \overline{T}$.

This shows that $\overline{S}' \in Min_b(SE(\Psi))$, where $(X', Y') \in SE(P_0)$ and $(X', Y') \in SE(P_2)$. Consequently, $SE(\Delta(\langle P_0, P_1, P_2\rangle) \sqcap P_2)$ is satisfiable.

($IC5'$). Consider $(X, Y) \in SE(\Delta(\langle P_0, \Psi\rangle) \sqcap \Delta(\langle P_0, \Psi'\rangle))$, and so $(X, Y) \in SE(\Delta(\langle P_0, \Psi\rangle))$ and $(X, Y) \in SE(\Delta(\langle P_0, \Psi'\rangle))$. Thus, $(X, Y) \in Min_b(SE(\langle P_0, \Psi\rangle))$ and $(X, Y) \in Min_b(SE(\langle P_0, \Psi'\rangle))$. Hence, there is some $\langle(X, Y), \overline{S}\rangle \in SE(\langle P_0, \Psi\rangle)$ and some $\langle(X, Y), \overline{S}'\rangle \in SE(\langle P_0, \Psi'\rangle)$ such that $\langle(X, Y), \overline{S}\rangle \leq_b \overline{T}$ for every $\overline{T} \in SE(\langle P_0, \Psi\rangle)$ and $\langle(X, Y), \overline{S}'\rangle \leq_b \overline{T}'$ for every $\overline{T}' \in SE(\langle P_0, \Psi'\rangle)$. But this implies that $\langle(X, Y), \overline{S}, \overline{S}'\rangle \leq_b \langle(X, Y), \overline{T}''\rangle$ for every $\overline{T}'' \in SE(\langle P_0, \Psi, \Psi'\rangle)$. Consequently, $(X, Y) \in SE(\Delta(\langle P_0, \Psi \circ \Psi'\rangle))$.

($IC7'$). If $\Delta(\langle P_0, \Psi\rangle) \sqcap P_1$ is unsatisfiable then the result is immediate.

So, assume that $\Delta(\langle P_0, \Psi\rangle) \sqcap P_1$ is satisfiable, and let $(X, Y) \in SE(\Delta(\langle P_0, \Psi\rangle) \sqcap P_1)$. That is, $(X, Y) \in SE(\Delta(\langle P_0, \Psi\rangle))$ and $(X, Y) \in SE(P_1)$. By definition we have that $(X, Y) \in Min_b(SE(\langle P_0, \Psi\rangle))_0$. Clearly, since $(X, Y) \in SE(P_1)$ we also obtain that $(X, Y) \in Min_b(SE(\langle P_0 \sqcap P_1, \Psi\rangle))_0$, from which we get $(X, Y) \in SE(\Delta(\langle P_0 \sqcap P_1, \Psi\rangle))$.

## A.10 Proof of Theorem 11

We first prove a helpful lemma.

LEMMA 5. *Let $\Psi$ be a belief profile. If $\overline{X} \in Min_a(SE(\Psi))$ then for $X_i \in \overline{X}$ we have $\langle X_i, \overline{X}\rangle \in Min_b(SE(\langle \emptyset, \Psi\rangle))$.*

PROOF. Let $\Psi$ be a belief profile, and let $\overline{X} \in Min_a(SE(\Psi))$. Hence, for every $\overline{Y} \in SE(\Psi)$ we have that $\overline{Y} \leq_a \overline{X}$ implies $\overline{X} \leq_a \overline{Y}$. Now, $\overline{Y} \leq_a \overline{X}$ means that $Y_i \ominus Y_j \subseteq X_i \ominus X_j$ for every $1 \leq i, j \leq |\Psi|$.

So, for fixed $i$ we have that $Y_i \ominus Y_j \subseteq X_i \ominus X_j$ implies that $X_i \ominus X_j \subseteq Y_i \ominus Y_j$. Let $X_0 = X_i$ for that $i$. Thus, substituting we get that $Y_i \ominus Y_j \subseteq X_0 \ominus X_j$ implies that $X_0 \ominus X_j \subseteq Y_i \ominus Y_j$.

But this means that $\langle X_0, X\rangle \in Min_b(SE(\langle \emptyset, \Psi\rangle))$. □

For the proof of the theorem, we have:

Let $X \in SE(\nabla(\Psi))$. Then, $X \in \bigcup Min_a(SE(\Psi))$; that is, there is some $\overline{X}$ such that $X \in \overline{X}$ and $\overline{X} \in Min_a(SE(\Psi))$. But by Lemma 5 we then have that $\langle X, \overline{X} \rangle \in Min_b(SE(\langle \emptyset, \Psi \rangle))$. Hence, $X \in Min_b(SE(\langle \emptyset, \Psi \rangle)_0)$ and so $X \in SE(\Delta(\langle \emptyset, \Psi \rangle))$.

## A.11   Proof of Theorem 12

These results follow directly from the appropriate definitions.